\documentclass[12pt,letterpaper,twoside]{article}

    \usepackage[british]{babel}
    \usepackage[utf8]{inputenc}
    \usepackage[T1]{fontenc}
    \usepackage{lmodern}
    \usepackage{indentfirst}

    \usepackage{multirow}
    \usepackage[
      includefoot,
      top=2.5cm,
      bottom=2.5cm,
      inner=2.6cm,  
      outer=2.3cm,
      heightrounded     
    ]{geometry}

    \usepackage{fancyhdr}
    \usepackage{lastpage} 
    \usepackage{mathtools}    
    \usepackage{amssymb}      
    \usepackage{amsthm}       
    \usepackage{mathrsfs}     

    \usepackage{algorithm}
    \usepackage{algpseudocode}
    \floatname{algorithm}{Algorithm}
    \algrenewcommand\algorithmicrequire{\textbf{Input:}}
    \algrenewcommand\algorithmicensure{\textbf{Output:}}
    \usepackage{bm}

    \usepackage{graphicx}
    \usepackage[font=small,labelfont=bf,labelsep=colon]{caption}
    \usepackage{subcaption} 
    \usepackage{float}

    \usepackage{microtype}
    \microtypesetup{expansion=true,tracking=true,protrusion=true,final}
    \UseMicrotypeSet[protrusion]{basicmath}

    \usepackage[table,dvipsnames]{xcolor}
    \usepackage{hyperref}
    \hypersetup{
        colorlinks,
        linkcolor=RoyalBlue,
        citecolor=MidnightBlue,
        urlcolor=BlueViolet,
        bookmarksopen=true,
        pdftitle={Efficient Transformer-Inspired Variants of Physics-Informed Deep Operator Networks},
        pdfauthor={Zhi-Feng Wei; Wenqian Chen; Panos Stinis},
        pdfkeywords={DeepONet, Transformer, Operator Learning, PDE, Physics-Informed Machine Learning},
        pdfsubject={Physics-Informed Operator Learning for Partial Differential Equations}
    }
    \usepackage{xurl}

    \usepackage{bookmark}

    \usepackage[
        backend = biber,
        style = philosophy-modern,
        isbn = false,
        square = true,
        hyperref=true,
        url = true,
        doi = true,
        parentracker=true,
        shorthandintro = true]{biblatex}  
    \AtEveryBibitem{%
        \clearfield{abstract}
        \clearfield{keywords}
        \clearfield{addendum}
    }

    \DeclareFieldFormat{abstract}{\par\mkbibacro{ABSTRACT}: \small#1}
    \DeclareFieldFormat{addendum}{\par\mkbibacro{Note}: \small#1\nopunct}
    
    \NewBibliographyString{keyword,keywords}
    \DefineBibliographyStrings{english}{%
      keyword  = {keyword},
      keywords = {keywords},
    }
    
    \newcounter{cbx@keyword@total}
    \newcounter{cbx@keyword@count}
    \newcommand*{\keywordscount}[1]{%
      \stepcounter{cbx@keyword@total}}
    
    \newcommand*{\keywordsprint}[1]{%
      \stepcounter{cbx@keyword@count}%
      \ifnumless{\value{cbx@keyword@count}}{2}
        {}
        {\addcomma\space}%
      #1}
    \DeclareFieldFormat{keywords}{%
      \setcounter{cbx@keyword@total}{0}%
      \setcounter{cbx@keyword@count}{0}%
      \forcsvlist{\keywordscount}{#1}%
      \ifnumgreater{\value{cbx@keyword@total}}{1}
        {\par\mkbibacro{keywords}}
        {\par\mkbibacro{keyword}}%
      \addcolon\space
      \forcsvlist{\keywordsprint}{#1}%
    }
    \renewbibmacro*{finentry}{\printfield{abstract}\printfield{keywords}\finentry} 
    
    \DeclareFieldFormat{doi}{%
    \newline
    \mkbibacro{DOI}\addcolon\space
    \ifhyperref
      {\href{https://doi.org/#1}{\nolinkurl{#1}}}
      {\nolinkurl{#1}}
    }

    \DeclareFieldFormat{eprint:arxiv}{%
      \newline
      \mkbibacro{arXiv}\addcolon\space
      \href{https://arxiv.org/abs/#1}{#1}%
      \iffieldundef{eprintclass}{}{\addspace\mkbibbrackets{\printfield{eprintclass}}}%
    }
    \DeclareFieldAlias{eprint:arXiv}{eprint:arxiv} 

    \DeclareFieldFormat{url}{%
    \newline
    \mkbibacro{url}\addcolon\space
    \ifhyperref
      {\href{#1}{\nolinkurl{#1}}}
      {\nolinkurl{#1}}
    }
    
    \renewbibmacro*{doi+eprint+url}{%
      \iffieldequalstr{eprinttype}{arXiv}
        {%
          \iftoggle{bbx:eprint}{\usebibmacro{eprint}}{}%
        }%
        {%
          \printfield{doi}%
          \newunit\newblock
          \iffieldundef{doi}{\usebibmacro{url+urldate}}{}%
        }%
    }

    \usepackage{titlecaps}
    \Addlcwords{on and for a is but with of in as the etc to if its de l' von}
    
    \DeclareFieldFormat[article,inbook,incollection,inproceedings,patent,thesis,unpublished,misc]{titlecase}{\MakeSentenceCase*{#1}}
    \DeclareFieldFormat[misc]{title}{\MakeSentenceCase*{#1}}
    \DeclareFieldFormat[phdthesis,book]{titlecase}{\itshape\titlecap{#1}}
    
    \DeclareFieldFormat{jtu}{\itshape\titlecap{#1}}
    \renewbibmacro*{journal}{%
      \iffieldundef{journaltitle}
        {}
        {\printtext[journaltitle]{%
          \printfield[jtu]{journaltitle}%
          \setunit{\subtitlepunct}%
          \printfield[jtu]{journalsubtitle}}}}
    \DeclareFieldFormat[inproceedings]{booktitle}{
        \printfield[jtu]{booktitle}%
          \setunit{\subtitlepunct}%
          \printfield[jtu]{booksubtitle}
        }

    \NewBibliographyString{artno}
    \DefineBibliographyStrings{english}{artno = {article}}
    \DeclareFieldFormat[article,periodical]{eid}{\bibstring{artno}\addabbrvspace #1}

    \DefineBibliographyStrings{english}{jourvol = {vol\adddot}}
    \DefineBibliographyStrings{english}{number = {issue}}

    \DeclareFieldFormat[article]{volume}{\bibstring{jourvol}\addnbspace #1}
    \DeclareFieldFormat[article]{number}{\bibstring{number}\addnbspace #1}
    \renewbibmacro*{volume+number+eid}{%
      \printfield{volume}%
      \setunit{\addcomma\space}
      \printfield{number}%
      \setunit{\addcomma\space}%
      \printfield{eid}}
    
    \usepackage{cleveref}
    \usepackage{appendix}
    
    \usepackage{cases}

    \theoremstyle{plain}

    \theoremstyle{definition}
    
    \theoremstyle{remark}
    
    \theoremstyle{remark}

    \usepackage{booktabs}

    \DeclareMathOperator{\pb}{\mathbf{P}\mathopen{}}

    \newcommand\Pleft[1]{\pb\mkern-1mu\left[#1\right]}

    \newcommand\Psub [1]{\pb_{\! #1}}
    \newcommand\Psubleft[2]{\Psub {#1}\mkern-1mu\left[#2\right]}

    \DeclarePairedDelimiter\abs{\lvert}{\rvert} 

    \newcommand{\RNS}{\mathbb{R}}
    
    \newcommand{\ud}{\mathrm{d}} 

    \usepackage{yfonts} 
    \usepackage{calligra} 

    \usepackage{datetime2}
    \newcommand{\USdate}{\DTMenglishmonthname{\month}\space\number\day, \number\year}
    \usepackage{fontawesome5} 
    \newcommand{\redheart}{\textcolor{red}{\faHeart}}
    \newcommand\tstamp{\thanks{Current version: \USdate. Typeset with \redheart\ and \LaTeX.}}
    
    \usepackage{titling} 
    \pretitle{\begin{center}\LARGE\textbf}
    \posttitle{\end{center}}
    \title{Efficient Transformer-Inspired Variants of Physics-Informed Deep Operator Networks\tstamp}

    \usepackage[affil-it]{authblk} 
    \newcommand\corref[2]{\thanks{Email: \href{mailto:#1}{#1}. #2}}

    \author[1]{Zhi-Feng Wei\corref{zfwei@pnnl.gov}{}}
    \author[1]{Wenqian Chen\corref{wenqian.chen@pnnl.gov}{}}
    \author[1,2,3]{Panos Stinis\corref{panagiotis.stinis@pnnl.gov}{}}

    \affil[1]{Advanced Computing, Mathematics, and Data Division,
        Pacific Northwest National Laboratory, Richland, WA 99354, USA}
    \affil[2]{Department of Applied Mathematics, University of Washington,
        Seattle, WA 98195, USA}
    \affil[3]{Division of Applied Mathematics, Brown University,
        Providence, RI 02912, USA}

    \predate{}
    \date{}
    \postdate{}

    \providecommand{\keywords}[1]{\flushleft\textbf{\textit{Keywords:}} #1}  
    
    \usepackage{tabularx}
    \newcolumntype{M}{>{$\displaystyle}X<{$}} 
    \DeclareRobustCommand{\tdeeponet}[1]{T\nobreakdash-DeepONet\nobreakdash--#1}
    \usepackage{pdflscape}

\begin{document}
\pagenumbering{roman}
\setcounter{page}{1}
\maketitle\thispagestyle{empty}
\begin{abstract}
Operator learning has emerged as a promising tool for accelerating the solution of partial differential equations (PDEs). The Deep Operator Networks (DeepONets) represent a pioneering framework in this area: the ``vanilla'' DeepONet is valued for its simplicity and efficiency, while the modified DeepONet achieves higher accuracy at the cost of increased training time.
In this work, we propose a series of \emph{Transformer-inspired} DeepONet variants that introduce bidirectional cross-conditioning between the branch and trunk networks in DeepONet. Query-point information is injected into the branch network and input-function information into the trunk network, enabling dynamic dependencies while preserving the simplicity and efficiency of the ``vanilla'' DeepONet in a non-intrusive manner.
Experiments on four PDE benchmarks---advection, diffusion--reaction, Burgers', and Korteweg--de Vries equations---show that for each case, there exists a variant that matches or surpasses the accuracy of the modified DeepONet while offering improved training efficiency. Moreover, the best-performing variant for each equation aligns naturally with the equation's underlying characteristics, suggesting that the effectiveness of cross-conditioning depends on the characteristics of the equation and its underlying physics.
To ensure robustness, we validate the effectiveness of our variants through a range of rigorous statistical analyses, among them the Wilcoxon Two One-Sided Test, Glass's Delta, and Spearman's rank correlation.


\end{abstract}
\clearpage
\setcounter{tocdepth}{2}
\tableofcontents
\thispagestyle{empty}
\clearpage
\pagenumbering{arabic}
\setcounter{page}{1}

\section{Introduction}
Partial differential equations (PDEs) play a central role in modeling physical, engineering, and biological phenomena. They govern a wide spectrum of processes, such as fluid dynamics, chemical reactions, and wave propagation \autocite{Evans-2022}. 
While classical numerical solvers for PDEs are accurate, they become computationally prohibitive in high-dimensional or parameterized settings \autocite{Quarteroni-2015,Hesthaven-2016,Han-2018,E-2018}. 
To overcome these challenges, \emph{operator learning} has emerged as a promising alternative \autocite{Boulle-2024,Lu-2021,Li-2020}. By training neural networks to approximate solution operators, this approach enables fast and flexible inference after training. Moreover, \emph{physics-informed training strategies} reduce reliance on large datasets and maintain performance in data-scarce regimes \autocite{Raissi-2019}, though improving their convergence remains an active area of research \autocite{Krishnapriyan-2021,Wang-2022-ntk,Chen-2025,Wang-2023,Howard-2025}.

As a pioneering operator learning framework, the Deep Operator Network (hereafter referred to as ``vanilla'' DeepONet) demonstrated the potential of neural operators for solving PDEs \autocite{Lu-2021,Wang-2021}. It consists of two distinct subnetworks: a \emph{branch network} that encodes the input function (e.g., the initial condition or source term) and a \emph{trunk network} that encodes the query spatiotemporal point. These two components operate independently during inference, which, while structurally elegant, can limit the model's ability to capture complex interactions between input function and query point \autocite{Kovachki-2023}.
In response to this architectural limitation, various extensions have been proposed to improve representational capacity, including SVD-DeepONet \autocite{Venturi-2023}, the R-adaptive DeepONet designed for discontinuities \autocite{Zhu-2024}, multi-input EDeepONet \autocite{Tan-2022}, and the Separable Operator Networks \autocite{Yu-2024}. These models reflect an active line of research focused on enhancing DeepONet's expressive capacity through architectural innovations.
Specifically, to improve the predictive accuracy of the ``vanilla'' DeepONet, a modified DeepONet architecture with increased structural complexity was proposed in prior work \autocite{Wang-2022}. While this design demonstrates improved predictive accuracy across a range of PDE benchmarks, it typically requires two to three times longer training time. The original ``vanilla'' DeepONet, on the other hand, remains attractive for its computational simplicity. These contrasting approaches reflect an important trade-off between accuracy and efficiency. Motivated by these developments, we aim to explore architectural enhancements that build on existing insights---drawing inspiration from the attention mechanism in Transformers---to improve \emph{accuracy} while preserving \emph{efficiency}.

The Transformer architecture has revolutionized sequence modeling in natural language processing by 
capturing long-range dependencies through the attention mechanism \autocite{Vaswani-2017}. This capacity to model both long-range and cross-input dependencies has also made attention mechanisms appealing for operator learning, where interactions between input functions and query spatiotemporal points are crucial. Motivated by this, several studies have introduced Transformers or attention-based modules into PDE operator learning, such as the operator transformer \autocite{Li-2022}, LOCA architecture with coupled attention \autocite{Kissas-2022}, Position-induced Transformer \autocite{Wu-2024}, the GNOT architecture for multiple input functions and irregular meshes \autocite{Hao-2023}, and the use of transformers for differential equations with finite regularity \autocite{Shih-2025}. These promising directions have significantly expanded the design space for operator learning. However, these approaches often come with increased architectural complexity, which may raise the cost of implementation and training in practice. In this work, we explore a complementary approach focused on non-intrusive dependency modeling, aiming to retain simplicity and efficiency.

Building on the observation that the ``vanilla'' DeepONet lacks \emph{cross-dependence between its two subnetworks}---the branch network depends solely on the input function, while the trunk network depends solely on the query spatiotemporal point---and that there is \emph{no communication} or exchange of information between them, we propose Transformer-inspired DeepONet variants. Motivated by attention mechanisms \autocite{Vaswani-2017,Phuong-2022}, we introduce query-point information into the branch network, enabling its output to vary dynamically with respect to the query point. Conversely, we provide the trunk network with information from the input function, allowing its outputs to better capture the underlying problem context. These modifications establish \emph{bidirectional interaction} between the branch and trunk networks. The design remains \emph{dependency-driven}, \emph{non-intrusive}, and \emph{architecturally simple}. As a result, the proposed variants retain the training efficiency of the ``vanilla'' DeepONet while achieving predictive accuracy comparable to or even exceeding that of the modified DeepONet, with minimal added complexity. We demonstrate the effectiveness of our approach through evaluations on four PDE benchmarks: advection, diffusion--reaction, Burgers', and Korteweg--de Vries equations. 
To ensure the robustness of our findings, we apply rigorous statistical tools---including the Wilcoxon Two One-Sided Test (TOST), Glass's Delta, and Spearman's rank correlation $\rho$---to compare the predictive accuracy of the proposed model variants.

The remainder of this paper is organized as follows. \Cref{sec_technical} presents the key technical background and describes the architecture of our Transformer-inspired DeepONet variants in detail. \Cref{sec_results} reports numerical results on four PDE benchmarks. \Cref{sec_discussion} offers discussion and concluding remarks. Finally, the appendices detail data generation procedures, model training settings, raw results, and the rigorous statistical analysis methods used to support our findings. 

\section{Technical Approach}\label{sec_technical}
    This section introduces the architectures of the operator learning models investigated in this study. We first describe the foundational ``vanilla'' DeepONet \autocite{Lu-2021,Wang-2021} and modified DeepONet \autocite{Wang-2022} frameworks in \Cref{sec_don}. Building upon these concepts, \Cref{sec_ti_don_var} then details the structure of our proposed Transformer-inspired DeepONet variants.

    We consider \emph{time-dependent} partial differential equations (PDEs), where certain properties of the system are allowed to vary. These properties, represented by input functions, can define the initial conditions or source terms. To formalize this problem, let $(\mathcal{U}, \mathcal{V}, \mathcal{S})$ be a triplet of Banach spaces. A general time-dependent PDE can be expressed with a linear or nonlinear operator $\mathcal{N}\colon\mathcal{U}\times\mathcal{S}\to\mathcal{V}$ as:
    \begin{equation}
        \bm{s}_t + \mathcal{N}(\bm{u}, \bm{s}) = 0, \label{pde_general}
    \end{equation}
    where $\bm{u} \in \mathcal{U}$ is the input function and $\bm{s} \in \mathcal{S}$ is the corresponding unknown solution. In the concrete PDEs considered later, $\bm{u}$ typically represents the initial condition, coefficient, or a source term; in such cases, $\bm{u}(\bm{x})$ and the solution $\bm{s}(t, \bm{x})$ share the same spatial domain. We assume that for any given $\bm{u} \in \mathcal{U}$, a unique solution $\bm{s}$ exists under appropriate initial and boundary conditions. The solution operator $G\colon \mathcal{U}\to\mathcal{S}$ is thus defined as
    \begin{equation}\label{sol_operator}
        G(\bm{u}) \coloneq \bm{s}.
    \end{equation}
    Here, the solution $\bm{s} = \bm{s}(t, \bm{x})$ is a function of both time $t$ and spatial coordinates $\bm{x}$, while the input function $\bm{u} = \bm{u}(\bm{x})$ is a function of the spatial coordinates only.
    
    \subsection{Physics-Informed DeepONet}\label{sec_don}
    
\begin{figure}[htbp]
            \centering
            \includegraphics[width=14cm]{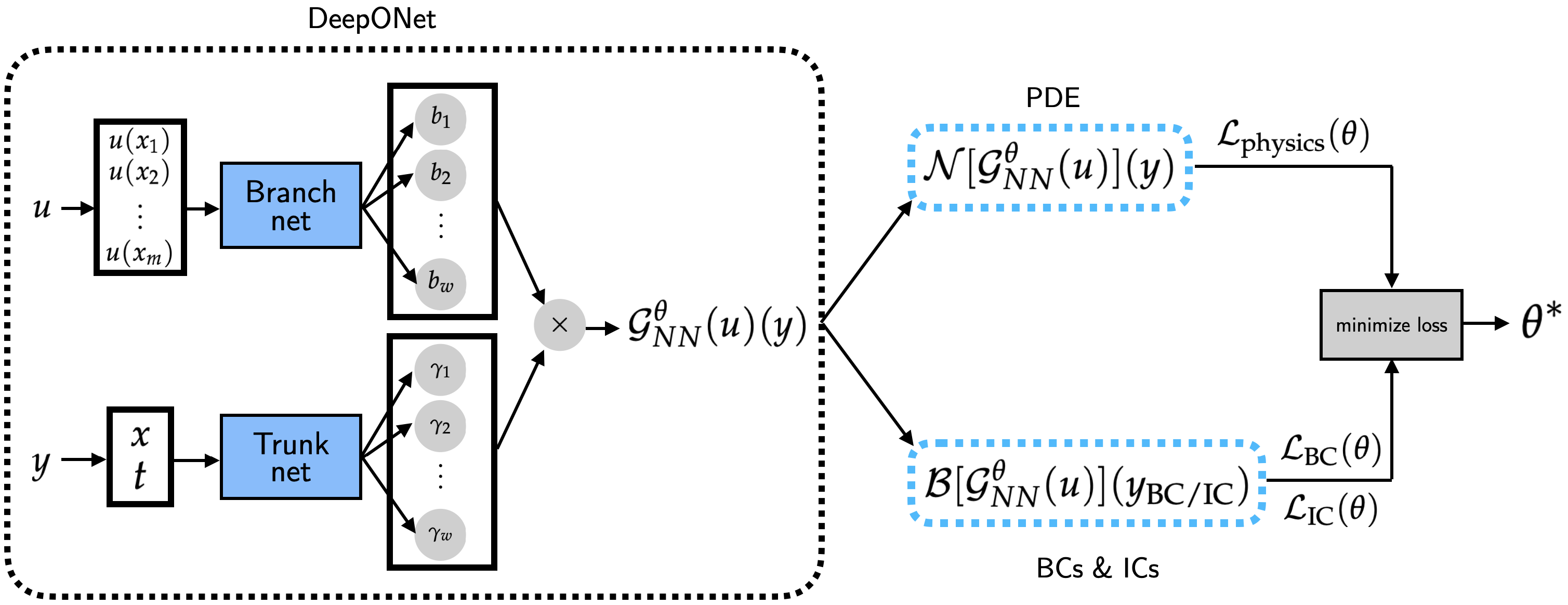}
            \caption{\label{fig_don_struct_vanilla}Physics-informed ``vanilla'' DeepONet (From \autocite{Williams-2024}).}
        \end{figure}

        To approximate the solution operator $G$, the ``vanilla'' DeepONet framework can be employed \autocite{Lu-2021}. As illustrated in \Cref{fig_don_struct_vanilla}, a ``vanilla'' DeepONet consists of two distinct neural networks: a ``branch network'' and a ``trunk network''.
        
        The branch net processes the input function $\bm{u}$. Specifically, it takes a discrete representation of $\bm{u}$, evaluated at a fixed set of sensor spatial locations $\{\bm{x}_i\}_{i=1}^m$, and produces a feature embedding $(b_1, b_2, \ldots, b_w)^\intercal \in \RNS^w$. The trunk net takes the continuous spatiotemporal coordinates $\bm{y}=(t, \bm{x})$ as input and generates a corresponding feature embedding $(\gamma_1, \gamma_2, \ldots, \gamma_w)^\intercal \in \RNS^w$.
        The final prediction is formed by computing the inner product of these two embeddings. The output of the ``vanilla'' DeepONet, representing the solution $\bm{s}$ for a given function $\bm{u}$,  evaluated at spatiotemporal coordinates $\bm{y}=(t, \bm{x})$, is expressed as:
        \begin{equation}
            G^{\bm{\theta}}(\bm{u})(\bm{y})
            = \sum_{k=1}^{w}
            \underbrace{b_{k}\bigl(\bm{u}(\bm{x}_{1}), \bm{u}(\bm{x}_{2}), \ldots, \bm{u}(\bm{x}_{m})\bigr)}_{\text{branch net}}
            \underbrace{\gamma_{k}(\bm{y})}_{\text{trunk net}},
        \end{equation}
        where $\bm{\theta}$ represents the collection of all trainable parameters in both branch net and trunk net.

The key insight behind \emph{physics-informed} ``vanilla'' DeepONets \autocite{Wang-2021} is that the model output is differentiable with respect to its spatiotemporal input coordinates $\bm{y}=(t, \bm{x})$. This property allows the use of automatic differentiation to compute the derivatives of the predicted solution $\bm{s}$, which can then be substituted into the governing PDE (\Cref{pde_general}) to construct a residual loss term. By minimizing this physics-based residual, together with the losses associated with the initial and boundary conditions, the network can be trained to satisfy the underlying physical laws without requiring labeled solution data. In physics-informed training, multiple loss terms are typically combined, and their relative weights must be tuned to achieve rapid and stable convergence. To address this loss-balancing challenge, we employ both a moderately localized conjugate kernel (CK) weighting strategy (see \autocite{Qadeer-2023,Wang-2022}) and the balanced-residual-decay-rate (BRDR) self-adaptive weighting scheme \autocite{Chen-2025} in our numerical experiments.

\begin{figure}[htbp]
            \centering
            \includegraphics[width=14cm]{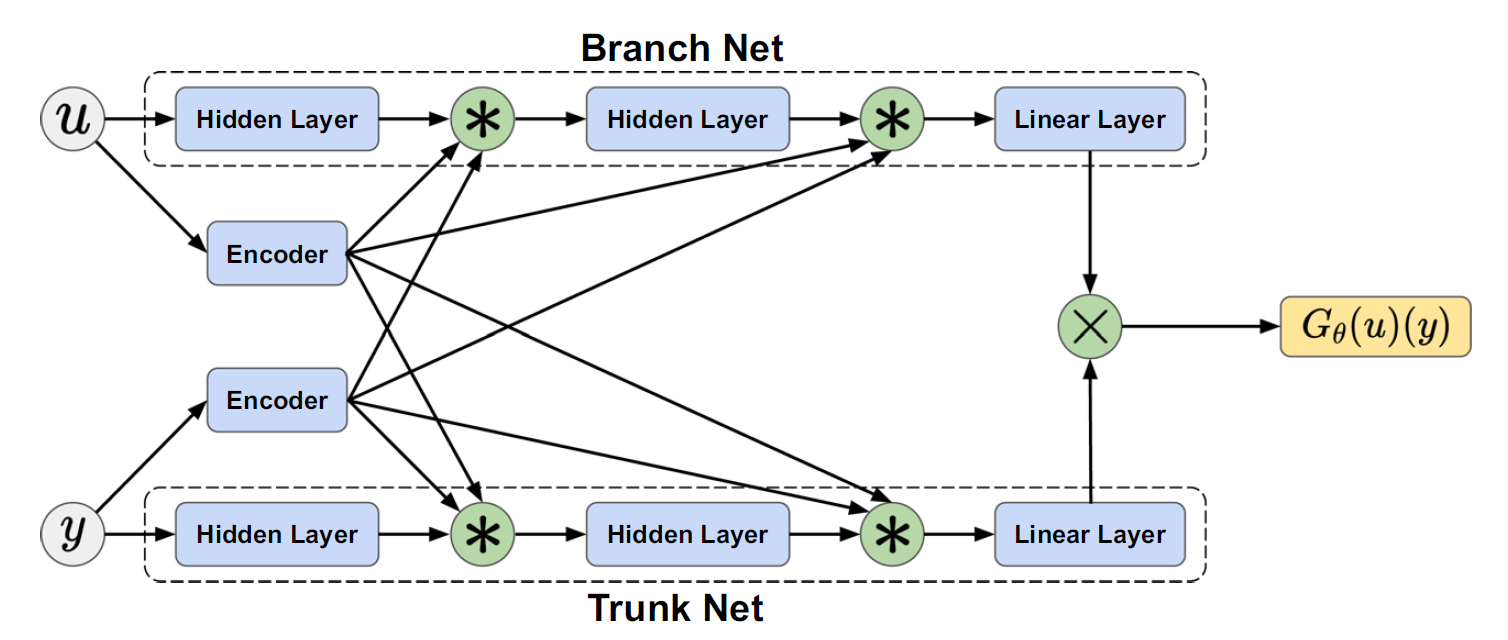}
            \caption{\label{fig_don_struct_modified}The modified DeepONet architecture (From \autocite{Wang-2022}).}
        \end{figure}

        In \autocite{Wang-2022}, an improved architecture called the ``modified DeepONet'' was proposed, as illustrated in \Cref{fig_don_struct_modified}. This architecture retains the ``vanilla'' DeepONet as a backbone but incorporates two additional encoders and modifies the forward pass. While these changes reportedly lead to more accurate predictions, they also incur longer training times due to increased structural complexity.

    \subsection{Transformer-Inspired DeepONet Variants}\label{sec_ti_don_var}

        The trade-off between predictive accuracy and computational cost inherent in the existing ``vanilla'' DeepONet and the modified DeepONet frameworks motivates our work. The modified DeepONet, as introduced in \autocite{Wang-2022}, exhibits superior predictive accuracy in solving PDEs. However, compared to the ``vanilla'' DeepONet, it imposes significantly higher training costs, particularly when dealing with higher-order PDEs, with training times often increasing by a factor of two or three (see comparison in \Cref{sec_results}). To address this challenge, we propose a series of Transformer-inspired variants built upon the ``vanilla'' DeepONet architecture. Our primary objective is to develop models that not only retain the predictive enhancements of the modified DeepONet but also preserve the training efficiency and structural simplicity inherent to the ``vanilla'' DeepONet framework.

        Before detailing our proposed models, we first briefly review the attention mechanism, the key component of the standard Transformer architecture, originally introduced in \autocite{Vaswani-2017}. Readers seeking a self-contained and mathematically precise overview of Transformer architectures and algorithms are referred to \autocite{Phuong-2022}.

        \subsubsection{The Transformer Self-Attention Mechanism}
        
        The main architectural component of the Transformer is the attention mechanism. \Cref{algo:basic_attention} provides the details of a basic single-query attention mechanism. In accordance with common mathematical conventions, the algorithm assumes that matrices act on column vectors placed to their right through left multiplication.
        
        \begin{algorithm}[htbp]
        \caption{Basic single-query attention (From \autocite{Phuong-2022}).}
        \label{algo:basic_attention}
        \begin{algorithmic}[1]
        \Require $\bm{e} \in \RNS^{d_{\text{in}}}$ \Comment{vector representation of the current token}
        \Require $\{\bm{e}_\tau\}_{\tau=1}^T \subseteq \RNS^{d_{\text{in}}}$ \Comment{vector representations of context tokens}
        \Ensure $\bm{\widetilde v} \in \RNS^{d_{\text{out}}}$ \Comment{attention output}
        
        \State \textbf{Parameters:}
        \State $\bm{W_q}, \bm{W_k} \in \RNS^{d_{\text{attn}} \times d_{\text{in}}},\quad \bm{b_q}, \bm{b_k} \in \RNS^{d_{\text{attn}}}$
        \State $\bm{W_v} \in \RNS^{d_{\text{out}} \times d_{\text{in}}},\quad \bm{b_v} \in \RNS^{d_{\text{out}}}$
        
        \State $\bm{q} \gets \bm{W_q} \bm{e} + \bm{b_q}$
        
        \For{$\tau = 1$ to $T$}
            \State $\bm{k}_\tau \gets \bm{W_k} \bm{e}_\tau + \bm{b_k}$
            \State $\bm{v}_\tau \gets \bm{W_v} \bm{e}_\tau + \bm{b_v}$
        \EndFor
        
        \For{$\tau = 1$ to $T$}
            \State 
            $\alpha_\tau
            \gets
            \dfrac{\exp\bigl(\bm{q}^\intercal \bm{k}_\tau / \sqrt{d_{\text{attn}}}\bigr)}
            { \sum_{\tau=1}^T \exp\bigl(\bm{q}^\intercal \bm{k}_\tau / \sqrt{d_{\text{attn}}} \bigr)}$
        \EndFor
        
        \State $\bm{\widetilde v} \gets \sum_{\tau=1}^T \alpha_\tau \bm{v}_\tau$
        \State \Return $\bm{\widetilde v}$
        \end{algorithmic}
        \end{algorithm}
        
        In this attention mechanism, the attention output $\bm{\widetilde{v}}$ for the current token $\bm{e}$ is computed as a weighted sum of the value vectors $\{\bm{v}_\tau\}_{\tau=1}^T$, each of which is obtained by applying a linear transformation to the corresponding context token $\bm{e}_\tau$. 
        The weights $\{\alpha_\tau\}_{\tau=1}^T$, known as attention scores, serve as the combination coefficients for the value vectors $\{\bm{v}_\tau\}_{\tau=1}^T$. As such, $\alpha_\tau$ directly determines the contribution of each context token to the final representation.
        These scores are computed via scaled dot-product, which measures the compatibility between the query vector $\bm{q}$ of the current token and the key vectors $\{\bm{k}_\tau\}_{\tau=1}^T$ of the context tokens.     
        
        \subsubsection{Relating Transformer Attention to DeepONet}
        
        By comparing the output formulations of the ``vanilla'' DeepONet and the attention mechanism (\Cref{algo:basic_attention}) used in Transformer, we can establish a conceptual correspondence. Both architectures compute an output as a linear combination of basis functions.
        
        In the DeepONet framework, the trunk net outputs, $\bigl\{\gamma_k(\bm{y})\bigr\}_{k=1}^w$, serve as the basis functions \autocite{Meuris-2023}, where $\bm{y}=(t, \bm{x})$ is the spatiotemporal coordinates. The corresponding coefficients, $\{b_k\}_{k=1}^w$, are generated by the branch net. The input to the branch net---the discrete function representation $\bigl\{\bm{u}(\bm{x}_i)\bigr\}_{i=1}^m$---can be viewed as the set of context tokens, as it provides the necessary context for the specific PDE instance. However, the resulting coefficients $\{b_k\}_{k=1}^w$ remain static with respect to $\bm{y} = (t, \bm{x})$ and do not dynamically adapt to specific query points.
        
        In contrast, the attention mechanism in Transformers introduces a dynamic dependency on the query point. Here, the value vectors $\{\bm{v}_\tau\}_{\tau=1}^T$, derived from linear transformations of the context tokens $\{\bm{e}_\tau\}_{\tau=1}^T$, act as basis functions. The coefficients $\{\alpha_\tau\}_{\tau=1}^T$ are computed dynamically based on the compatibility between the query vector $\bm{q}$ of the current token $\bm{e}$ and the key vectors $\{\bm{k}_\tau\}_{\tau=1}^T$, enabling the model to adapt to specific query points.

        This comparison reveals two fundamental distinctions:
        \begin{enumerate}
            \item \textbf{Query-Dependent Coefficients:} Transformers compute the coefficients $\{\alpha_\tau\}_{\tau=1}^T$ dynamically, allowing them to adjust based on the query vector $\bm{q}$. In the ``vanilla'' DeepONet framework, the coefficients $\{b_k\}_{k=1}^w$ remain static for a given input $\bm{u}$ and cannot adapt to variations in the query point $\bm{y} = (t, \bm{x})$.
            \item \textbf{Role of the Input ``Context'' Function:} In Transformers, the context tokens $\{\bm{e}_\tau\}_{\tau=1}^T$ interact with both the basis functions $\{\bm{v}_\tau\}_{\tau=1}^T$ and the coefficients $\{\alpha_\tau\}_{\tau=1}^T$, creating a bidirectional dependency. In contrast, in the ``vanilla'' DeepONet framework, the input ``context'' function $\bm{u}$ solely determines the coefficients $\{b_k\}_{k=1}^w$, without any direct influence on the basis functions $\bigl\{\gamma_k(\bm{y})\bigr\}_{k=1}^w$.
        \end{enumerate}
        
        These observations underscore fundamental limitations in the expressive power of the ``vanilla'' DeepONet architecture. Specifically, the static nature of its query-independent coefficients and the rigid separation of functional roles between the branch and trunk nets restrict its capacity to effectively model complex operator mappings with dynamic dependencies. To overcome these challenges, our proposed Transformer-inspired variants introduce dynamic query dependence and expand the influence of input ``context'' functions, establishing a dynamic coupling between basis functions and coefficients. 
        
        \subsubsection{Proposed Dependency-Enhanced Architectures}
        
        Based on the analysis above, we introduce a series of architectural variants that enhance the ``vanilla'' DeepONet framework by incorporating query-dependent coefficients that enable branch net computations to dynamically adjust to specific spatiotemporal query points, 
        along with trunk-side conditioning on the input ``context'' function to directly influence the basis functions.  
        Recall that, for time-dependent PDEs considered in this study, the input function $\bm{u}(\bm{x})$ and the corresponding solution $\bm{s}(t, \bm{x})$ share the same spatial domain.
                
        \paragraph{\texorpdfstring{\tdeeponet{Bx} (Query-Dependent Branch Net)}{T-DeepONet--Bx (Query-Dependent Branch Net)}.}
        To address the first limitation in the ``vanilla'' DeepONet architecture, we introduce a variant by injecting the spatial coordinate $\bm{x}$ as an additional input to the branch net. This modification enables the coefficients of the basis functions to dynamically depend on query points, reflecting the adaptive nature of attention scores in a Transformer. As such, this variant, referred to as \tdeeponet{Bx} (Bx), incorporates $\bm{x}$ directly into the branch net. The branch net computes the query-dependent coefficients $\bm{b} = \bm{b}(\bm{u}, \bm{x})$, and the model output is formulated as:
        \begin{equation}
            G^{\bm{\theta}}(\bm{u})(t, \bm{x}) = \sum_{k=1}^{w} b_{k}(\bm{u}, \bm{x})\gamma_{k}(t, \bm{x}).
        \end{equation}

        \paragraph{\texorpdfstring{\tdeeponet{TL} (Context-Aware Trunk Net)}{T-DeepONet--TL (Context-Aware Trunk Net)}.}  
        To address the second limitation in the ``vanilla'' DeepONet architecture, this variant introduces dependency of the trunk net on the input ``context'' function $\bm{u}$. Similar to how value vectors $\{\bm{v}_\tau\}_{\tau=1}^T$ in a Transformer are influenced by context tokens $\{\bm{e}_\tau\}_{\tau=1}^T$, the trunk net in \tdeeponet{TL} takes both the query point $(t, \bm{x})$ and the value of the input function at the specific spatial coordinate, $\bm{u}(\bm{x})$, as inputs. Consequently, the basis functions are defined as  
        $\bm{\gamma} = \bm{\gamma}\bigl(t, \bm{x}, \bm{u}(\bm{x})\bigr)$.  
        This variant is referred to as \tdeeponet{TL} (TL), emphasizing that the trunk net conditions on the local value $\bm{u}(\bm{x})$ to enhance its representation capabilities. The model output is expressed as:  
        \begin{equation}
            G^{\bm{\theta}}(\bm{u})(t, \bm{x})
            =
            \sum_{k=1}^{w} b_{k}(\bm{u})\gamma_{k}\bigl(t, \bm{x}, \bm{u}(\bm{x})\bigr).
        \end{equation}

        \paragraph{\texorpdfstring{\tdeeponet{BxTL} (Combined Local Context)}{T-DeepONet--BxTL (Combined Local Context)}.}  
        This variant integrates the principles of the first two architectures. Specifically, the branch net exhibits query-dependency, dynamically adapting its coefficients to the spatial input $\bm{x}$, while the trunk net introduces context-awareness by conditioning on the local value $\bm{u}(\bm{x})$ of the input function. We refer to this variant as \tdeeponet{BxTL} (BxTL), denoting branch-side injection of $\bm{x}$ and trunk-side local conditioning. The final model output is formulated as:  
        \begin{equation}
            G^{\bm{\theta}}(\bm{u})(t, \bm{x})
            =
            \sum_{k=1}^{w} b_{k}(\bm{u}, \bm{x})\gamma_{k}\bigl(t, \bm{x}, \bm{u}(\bm{x})\bigr).
        \end{equation}
        
        \paragraph{\texorpdfstring{\tdeeponet{BxTG} (Combined Global Context)}{T-DeepONet--BxTG (Combined Global Context)}.}  
        Building on the architecture of \tdeeponet{BxTL}, this variant introduces a more general framework by allowing the trunk net to access the entire input function $\bm{u}$. This modification equips the trunk net with global context when generating the basis functions, enhancing its ability to capture dependencies across the spatial domain. We refer to this variant as \tdeeponet{BxTG} (BxTG), highlighting the use of global context rather than restricting to the local value $u(\bm{x})$. The model output is expressed as:  
        \begin{equation}
            G^{\bm{\theta}}(\bm{u})(t, \bm{x})
            =
            \sum_{k=1}^{w} b_{k}(\bm{u}, \bm{x})\gamma_{k}(t, \bm{x}, \bm{u}).
        \end{equation}
                
        In this formulation, both the basis functions and their coefficients are directly influenced by the global context of the input function $\bm{u}$ and the specific query point $(t, \bm{x})$. This global integration allows the trunk and branch nets to share richer contextual information, facilitating enhanced expressiveness in modeling complex operator mappings. This approach aligns with the philosophy of the modified DeepONet architecture \autocite{Wang-2022}, which similarly promotes information exchange between the branch net and the trunk net. However, \tdeeponet{BxTG} achieves this through a straightforward implementation of dependency mechanisms inspired by the dynamic interactions in Transformer attention, enabling simultaneous integration of global context and query-specific features.
        
        \paragraph{\texorpdfstring{\tdeeponet{TF} (Context-Aware Trunk Net with Fourier Coefficients)}{T-DeepONet--TF (Context-Aware Trunk Net with Fourier Coefficients)}.}  
        For PDEs with periodic boundary conditions, where the input function $\bm{u}$ (as the initial condition) is often periodic, we leverage a compact representation of $\bm{u}$ using its truncated Fourier coefficients $\widehat{\bm{u}}_{\Lambda} \coloneqq (\hat{u}_{\bm{k}})_{\bm{k} \in \Lambda}$. Here, $\Lambda$ represents a low-frequency index set, which will be specified for each equation. These coefficients provide a global frequency-based summary of $\bm{u}$.  
        
        Building on \tdeeponet{TL}, this variant incorporates these Fourier coefficients directly into the trunk net. The trunk net takes the query point $(t, \bm{x})$ and the Fourier coefficients $\widehat{\bm{u}}_{\Lambda}$ as input to generate the basis functions, formulated as  
        $\bm{\gamma} = \bm{\gamma}(t, \bm{x}, \widehat{\bm{u}}_{\Lambda})$.  
        We refer to this variant as \tdeeponet{TF} (TF), highlighting the trunk's ability to utilize Fourier-based global context. The model output is expressed as:  
        \begin{equation}
            G^{\bm{\theta}}(\bm{u})(t, \bm{x})
            =
            \sum_{k=1}^{w} b_{k}(\bm{u})\gamma_{k}(t, \bm{x}, \widehat{\bm{u}}_{\Lambda}).
        \end{equation}  
        This design equips the trunk net with global frequency information about the input function, making it particularly effective for capturing the behavior of solutions subject to periodic boundary conditions.
                        
        \paragraph{\texorpdfstring{\tdeeponet{BxTF} (Combined Global Context with Fourier Coefficients)}{T-DeepONet--BxTF (Combined Global Context with Fourier Coefficients)}.}  
        Extending the principles of \tdeeponet{TF}, this variant builds upon \tdeeponet{BxTL} by incorporating a truncated set of Fourier coefficients $\widehat{\bm{u}}_{\Lambda}$ to represent the global context, rather than relying solely on the local context $\bm{u}(\bm{x})$. In this architecture, the branch net computes the coefficients $\{b_k\}_{k=1}^w$, which depend on the input function $\bm{u}$ and the spatial query coordinate $\bm{x}$. Simultaneously, the trunk net generates basis functions informed by the global frequency content encoded in the Fourier coefficients $\widehat{\bm{u}}_{\Lambda}$.  
        
        This variant, referred to as \tdeeponet{BxTF} (BxTF), highlights the integration of branch-side spatial information with trunk-side frequency-aware conditioning. The model output is expressed as:  
        \begin{equation}
            G^{\bm{\theta}}(\bm{u})(t, \bm{x})
            =
            \sum_{k=1}^{w} b_{k}(\bm{u}, \bm{x})\gamma_{k}(t, \bm{x}, \widehat{\bm{u}}_{\Lambda}).
        \end{equation}  
        This design allows the model to dynamically adjust the weights of its frequency-aware basis functions based on spatial query points, thereby enhancing its flexibility to effectively handle PDEs with periodic boundary conditions.
        
        \begin{table}[H]
        \centering
        \caption{Summary of Architectural Variants and Their Input Features. The second and third columns list the feature sets
        fed to the branch net and trunk net, respectively. Notation: $\bm u$ denotes the input ``context'' function (sampled at sensor points); $t\in\mathbb{R}$ is the temporal coordinate and $\bm x\in\mathbb{R}^d$ the spatial coordinates; $\bm u(\bm x)$ is the value of $\bm u$ at spatial location $\bm x$; $\widehat{\bm u}_{\Lambda}$ denotes the truncated Fourier coefficients.}
        \label{tab:variants}
        \setlength{\tabcolsep}{9pt}
        \renewcommand{\arraystretch}{1.25}
        \rowcolors{2}{gray!10}{}
        \begin{tabularx}{\linewidth}{l M M}
        \toprule
        \textbf{Variant} & \textbf{Branch input} & \textbf{Trunk input} \\
        \midrule
        ``vanilla'' DeepONet    & \bm u                     & t,\ \bm x \\
        modified DeepONet       & \bm u                     & t,\ \bm x \\
        \tdeeponet{Bx}          & \bm u,\ \bm x             & t,\ \bm x \\
        \tdeeponet{TL}          & \bm u                     & t,\ \bm x,\ \bm u(\bm x) \\
        \tdeeponet{BxTL}        & \bm u,\ \bm x             & t,\ \bm x,\ \bm u(\bm x) \\
        \tdeeponet{BxTG}        & \bm u,\ \bm x             & t,\ \bm x,\ \bm u \\
        \tdeeponet{TF}          & \bm u                     & t,\ \bm x,\ \widehat{\bm u}_{\Lambda} \\
        \tdeeponet{BxTF}        & \bm u,\ \bm x             & t,\ \bm x,\ \widehat{\bm u}_{\Lambda} \\
        \bottomrule
        \end{tabularx}
        \end{table}
        
        A summary of our Transformer-inspired DeepONet variants is presented in \Cref{tab:variants}. These designs are intentionally kept \emph{dependency-driven}, \emph{non-intrusive}, and \emph{architecturally simple}, ensuring that they preserve the training efficiency of the original ``vanilla'' DeepONet while introducing only minimal additional complexity.

        Having outlined the theoretical motivations and architectural details of the proposed DeepONet variants, we now shift our focus to their empirical evaluation. In the next section, we will conduct a comprehensive performance comparison against both the ``vanilla'' DeepONet and the modified DeepONet models, highlighting the advancements in accuracy and computational efficiency achieved by our variants across multiple benchmark problems. 

\section{Numerical Results}\label{sec_results}
In this section, we present numerical results for four PDEs: the advection equation, the diffusion--reaction equation, Burgers' equation with varying viscosity, and the Korteweg--de Vries equation. We compare the performance of the ``vanilla'' DeepONet, the modified DeepONet, and the Transformer-inspired variants described in \Cref{sec_technical}. The experiments assess both predictive accuracy and training efficiency, using rigorous statistical methods outlined in \Cref{apdx3_data_analysis}.

The experimental setup follows prior work: \autocite{Wang-2022} for the advection and Burgers' equations, \autocite{Wang-2021} for the diffusion--reaction equation, and \autocite{Williams-2024} for the KdV equation. For each PDE, we adopt whichever loss-weighting scheme (CK or BRDR) yields better accuracy, and also test deterministic and random Fourier feature embeddings. Reported metrics are averaged over multiple runs with different random seeds to mitigate stochastic variability. 
Further implementation details are provided in \Cref{apdx1_details}, while the raw data underlying figures, prediction--reference visualizations, and selected literature comparisons are collected in \Cref{apdx21_raw_data,apdx22_pred_comparison,apdx23_literature_results}.

\subsection{Advection Equation}

We first consider a one-dimensional linear advection equation with spatially varying coefficients, given by
\begin{equation}
    \frac{\partial s}{\partial t} + u(x) \frac{\partial s}{\partial x} = 0, 
    \quad (t, x) \in (0,1) \times (0,1),
\end{equation}
subject to the initial condition
\begin{equation}
    s(0, x) = f(x), \quad x \in [0, 1],
\end{equation}
and the boundary condition
\begin{equation}
    s(t, 0) = g(t), \quad t \in [0, 1],
\end{equation}
where $f(x) = \sin \pi x$ and $g(t) = \sin \frac{\pi t}{2}$. 
To ensure numerical stability, the advection velocity $u(x)$ is enforced to be strictly positive by defining
\begin{equation}
    u(x) = v(x) - \min_{x} v(x) + 1,
\end{equation}
where $v(x)$ is drawn from a Gaussian random field (GRF) with correlation length $l = 0.2$. 
The learning objective is to approximate the solution operator $G$ that maps the spatially varying coefficient $u(x)$ to the corresponding spatiotemporal solution $s(t, x)$.

\begin{figure}[H]
  \centering
  \includegraphics[width=0.54\textwidth]{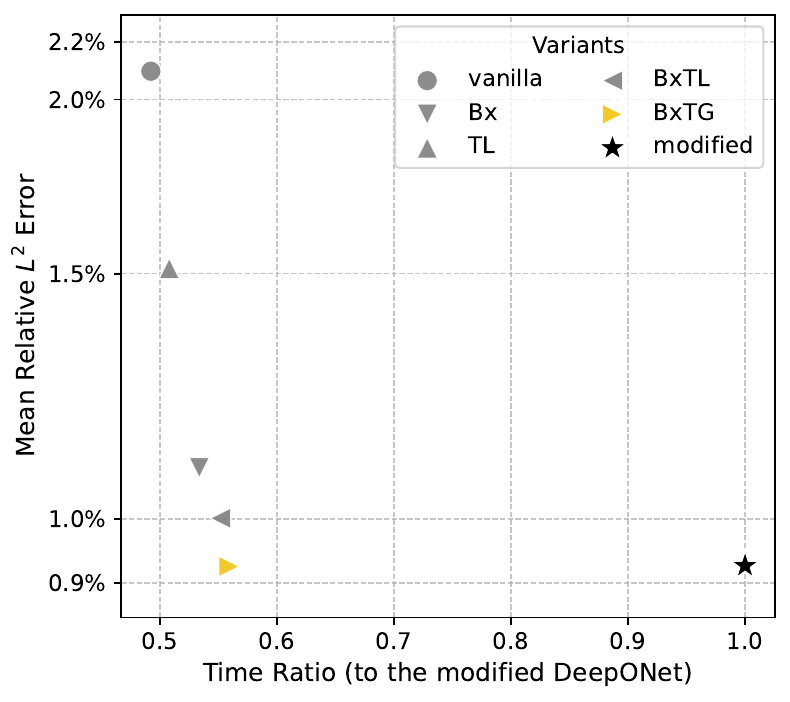}
  \caption{Performance comparison of model variants for the Advection equation. The horizontal axis shows the average per-iteration training time ratio relative to the modified DeepONet, averaged over multiple random seeds (lower is faster), while the vertical axis shows the mean relative $L^2$ error (lower is better). Marker shapes distinguish different model variants; the modified DeepONet is shown in black, and the best-performing variant is highlighted in gold.}
  \label{fig_advection_comparison}
\end{figure}

\begin{figure}[H]
  \centering
  \includegraphics[width=0.54\textwidth]{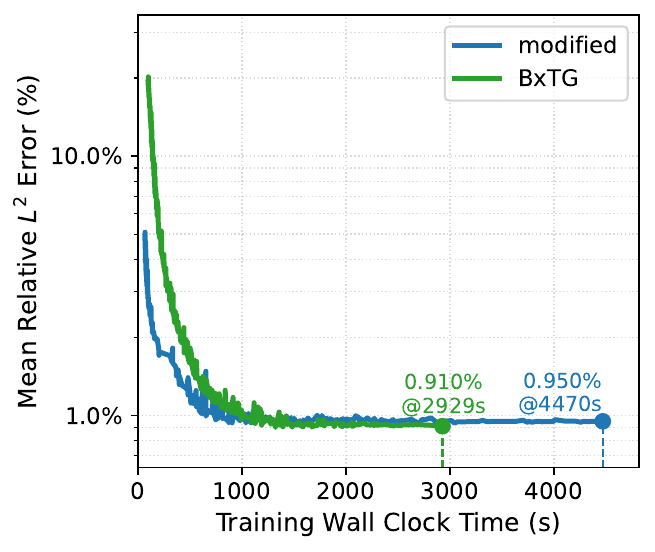}
  \caption{Comparison of training efficiency for the Advection equation. The horizontal axis shows the wall-clock training time for a single representative run, and the vertical axis shows the mean relative $L^2$ error over all test cases (log scale). Dashed vertical lines and annotations mark the time at which each model completed 300{,}000 training iterations.}
  \label{fig_advection_timecurve}
\end{figure}

\begin{figure}[H]
  \centering
  \includegraphics[width=0.54\textwidth]{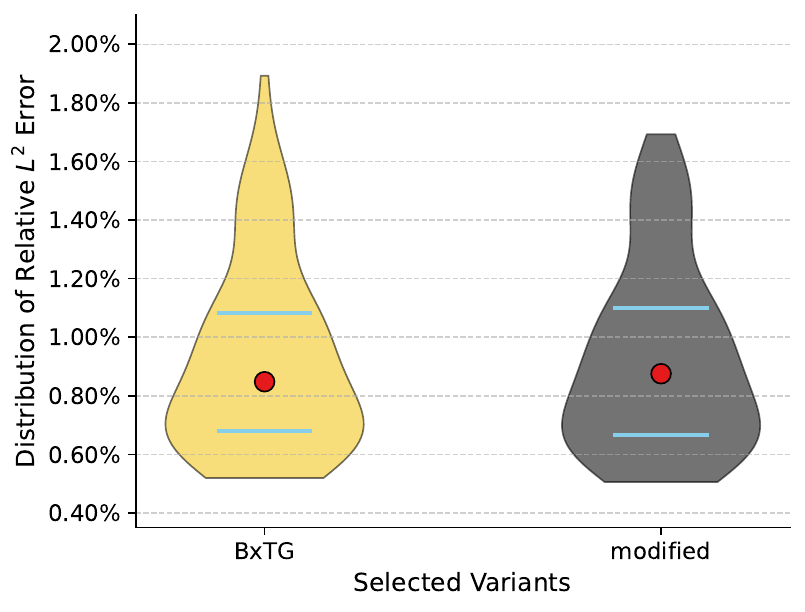}
  \caption{Distribution of relative $L^2$ errors for selected variants (gold: Variant BxTG; black: the modified DeepONet) on the Advection equation, evaluated over multiple random seeds and test instances. Horizontal bars span the interquartile range (25th to 75th percentile), and red dots mark the median.}
  \label{fig_advection_violin}
\end{figure}

\medskip
The performance of all evaluated models is summarized in \Cref{fig_advection_comparison,fig_advection_timecurve,fig_advection_violin}. 
The modified DeepONet (black star marker in~\Cref{fig_advection_comparison}) serves as the reference for both training time and accuracy. 
Among the Transformer-inspired variants, Variant BxTG (gold marker in~\Cref{fig_advection_comparison}) achieves the lowest mean relative $L^2$ error among the faster models, offering a speedup of nearly $45\%$ in average per-iteration training time over multiple random seeds, while maintaining comparable accuracy (mean relative $L^2$ error $0.925\%$ for BxTG and $0.926\%$ for the modified DeepONet). 
As an illustrative example, the convergence curves for one training run in \Cref{fig_advection_timecurve} show that Variant BxTG completes 300{,}000 training iterations in \( 2,929 \,\mathrm{s} \), compared to \( 4,470 \,\mathrm{s} \) for the modified DeepONet, representing a substantial reduction of about $35\%$ in the total time to complete the full training budget and a similar prediction accuracy.
However, the time required for the mean relative $L^2$ error to stabilize near its optimal value is only marginally shorter for Variant BxTG in this case, in contrast to other PDE problems discussed later (e.g., Burgers' equation), where both total training time and stabilization time show pronounced improvements. 
The strong performance of Variant BxTG can likely be attributed to its architectural modification that incorporates the full velocity field $u(x)$ directly into the trunk network input. In advection-dominated problems, the solution at each spatiotemporal location is largely determined by the transport of initial information along characteristic curves governed by the ordinary differential equation
$\tfrac{\ud x}{\ud t} = u(x).$
Providing the entire profile of $u(x)$---rather than only its local value at the evaluation point---allows the model to capture long-range dependencies and better infer the characteristic trajectories that underlie the transport behavior. This global encoding of the velocity field aligns naturally with the hyperbolic nature of the advection equation and enhances the model's ability to resolve directional propagation patterns, thereby contributing to improved predictive accuracy.
Additionally, the relatively simple architecture of our variants contributes to shorter training times. 
Error distributions over multiple test instances (\Cref{fig_advection_violin}) show that Variant BxTG has a slightly lower median relative $L^2$ error of $0.85\%$ than $0.88\%$ for the modified DeepONet, with similar spread.
In test cases, Variant BxTG achieved better predictive accuracy in $57.00\%$ of cases compared to the modified DeepONet.
A non-parametric Wilcoxon two one-sided tests (TOST) for equivalence (see \Cref{apdx3_data_analysis}), with a margin of $\pm 0.1\%$, confirms that Variant BxTG and the modified DeepONet are statistically equivalent in predictive accuracy, while a high Spearman correlation coefficient ($\rho = 0.993$) indicates strong agreement in error patterns. 

\paragraph{Overall:} Variant BxTG delivers predictive accuracy statistically equivalent to that of the modified DeepONet, while achieving a substantial speedup in training time.

\subsection{Diffusion--Reaction Equation}

We next consider a nonlinear diffusion--reaction equation with a spatially varying source term $u(x)$, formulated as
\begin{equation}
    \frac{\partial s}{\partial t} 
    = D \frac{\partial^{2} s}{\partial x^{2}} + k\, s^{2} + u(x), 
    \quad (t,x) \in (0,1] \times (0,1],
\end{equation}
subject to zero initial and boundary conditions:
\begin{align}
    s(0, x) &= 0, \quad x \in [0,1], \\
    s(t, 0) &= 0, \quad s(t, 1) = 0, \quad t \in [0,1].
\end{align}
Here, $D = 0.01$ denotes the diffusion coefficient and $k = 0.01$ the reaction rate. 
The objective is to learn the solution operator $G$ that maps the source term $u(x)$ to the corresponding spatiotemporal solution $s(t, x)$.

\begin{figure}[H]
  \centering
  \includegraphics[width=0.54\textwidth]{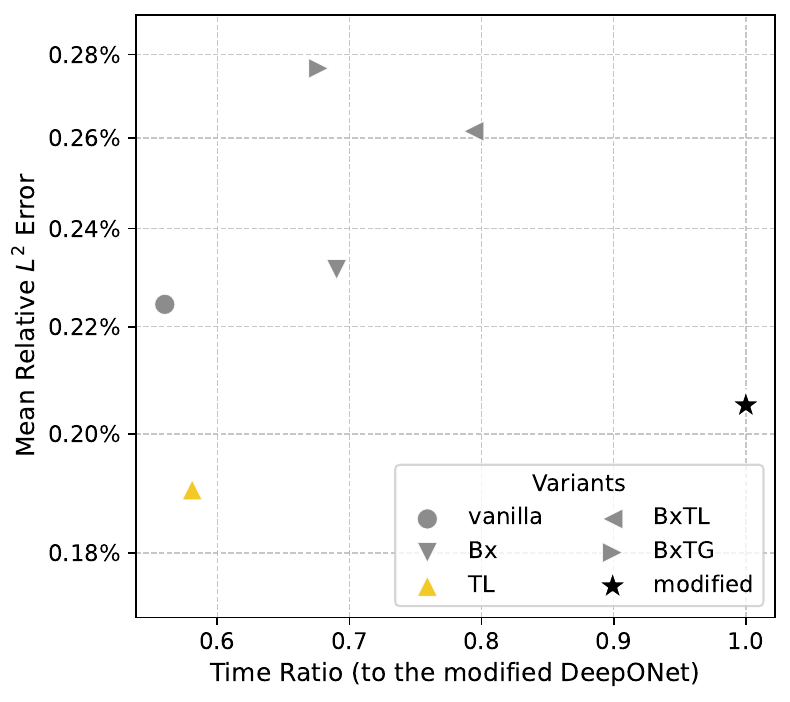}
  \caption{Performance comparison of model variants for the Diffusion--Reaction equation. The horizontal axis shows the average per-iteration training time ratio relative to the modified DeepONet, averaged over multiple random seeds (lower is faster), while the vertical axis shows the mean relative $L^2$ error (lower is better). Marker shapes distinguish different model variants; the modified DeepONet is shown in black, and the best-performing variant is highlighted in gold.}
  \label{fig_dre_comparison}
\end{figure}

\begin{figure}[H]
  \centering
  \includegraphics[width=0.58\textwidth]{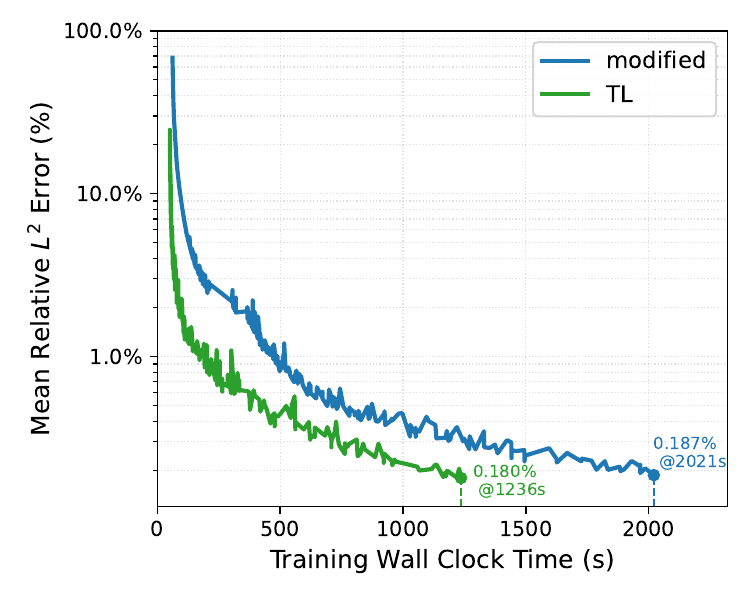}
  \caption{Comparison of training efficiency for the Diffusion--Reaction equation. The horizontal axis shows the wall-clock training time for a single representative run, and the vertical axis shows the mean relative $L^2$ error over all test cases (log scale). Dashed vertical lines and annotations mark the time at which each model completed 120{,}000 training iterations.}
  \label{fig_dre_timecurve}
\end{figure}

\begin{figure}[H]
  \centering
  \includegraphics[width=0.58\textwidth]{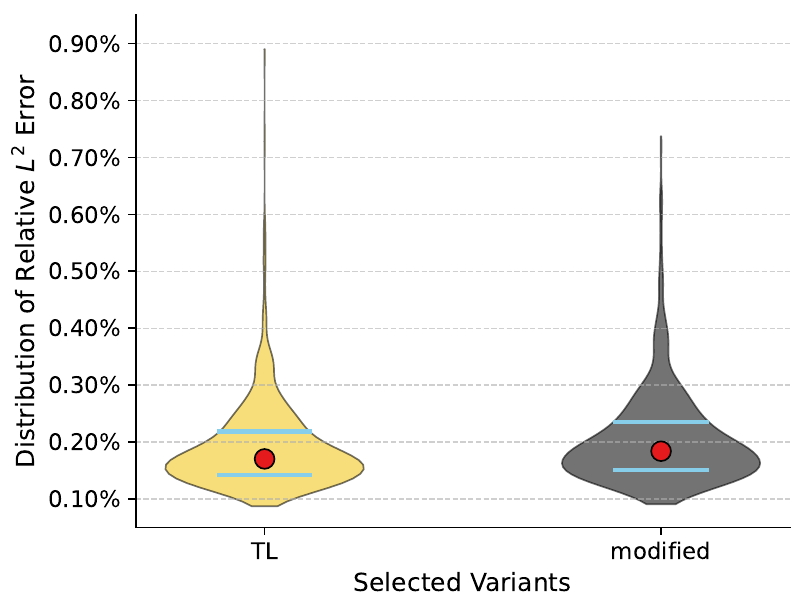}
  \caption{Distribution of relative $L^2$ errors for selected variants (gold: Variant TL; black: the modified DeepONet) on the Diffusion--Reaction equation, evaluated over multiple random seeds and test instances. Horizontal bars span the interquartile range (25th to 75th percentile), and red dots mark the median.}
  \label{fig_dre_violin}
\end{figure}

\medskip
The performance of all evaluated models is summarized in \Cref{fig_dre_comparison,fig_dre_timecurve,fig_dre_violin}. 
The modified DeepONet (black star marker in~\Cref{fig_dre_comparison}) serves as the reference for both training time and accuracy. 
Among the Transformer-inspired variants, Variant TL (gold marker in~\Cref{fig_dre_comparison}) achieves the lowest mean relative $L^2$ error among the faster models, offering an average per-iteration speedup of about $42\%$ over multiple random seeds, while maintaining comparable accuracy as the modified DeepONet. 
As an illustrative example, the convergence curves for one training run in \Cref{fig_dre_timecurve} show that Variant TL completes 120{,}000 training iterations in \( 1,236 \,\mathrm{s} \), compared to \( 2,021 \,\mathrm{s} \) for the modified DeepONet, representing a substantial reduction of about $39\%$ in the total time to complete the full training budget, alongside slightly improved accuracy. 
Moreover, at any given wall-clock time, the mean relative $L^2$ error of Variant TL (averaged over all test instances) remains consistently lower than that of the modified DeepONet, indicating faster convergence toward the optimum.
The strong performance of Variant TL is likely attributable to its architectural simplification compared to the modified DeepONet, as well as the direct inclusion of the local value $u(x)$ of the source term into the trunk network, which appears explicitly in the PDE. 
By leveraging this structural prior, the model can more effectively capture the dominant dependencies in the solution, enabling faster convergence and robust accuracy. 
Error distributions over all test instances (\Cref{fig_dre_violin}) show that Variant TL has a median relative $L^2$ error $0.17\%$, similar to the median relative $L^2$ error $0.18\%$ of the modified DeepONet. 
In test cases, Variant TL achieved better predictive accuracy in $64.90\%$ of cases compared to the modified DeepONet.
A non-parametric Wilcoxon two one-sided tests (TOST) for equivalence, with a margin of $\pm 0.02\%$, confirms that the two models are statistically equivalent in predictive accuracy, while a Spearman correlation coefficient $\rho = 0.769$ indicates strong agreement in error patterns.

\paragraph{Overall:} Variant TL delivers predictive accuracy that is statistically equivalent to that of the modified DeepONet, while substantially reducing training time.

\subsection{Burgers' Equation}

We next consider the one-dimensional viscous Burgers' equation,
\begin{equation}
    \frac{\partial s}{\partial t} 
    + s\,\frac{\partial s}{\partial x} 
    - \nu\,\frac{\partial^2 s}{\partial x^2} = 0, 
    \quad (t,x) \in (0,1] \times (0,1),
\end{equation}
subject to periodic boundary conditions
\begin{align}
    s(t, 0) &= s(t, 1), \\
    \frac{\partial s}{\partial x}(t, 0) &= \frac{\partial s}{\partial x}(t, 1),
\end{align}
and the initial condition
\begin{equation}
    s(0, x) = u(x), \quad x \in (0,1).
\end{equation}
We consider three viscosity values
\begin{displaymath}
\nu \in \{10^{-2},\, 10^{-3},\, 10^{-4}\}.
\end{displaymath}
The initial conditions $u(x)$ are sampled from a Gaussian random field (GRF)
\begin{displaymath}
u \sim \mathcal{N}\bigl(0,\, 25^2(-\Delta+5^2 I)^{-4}\bigr)
\end{displaymath}
that satisfies the periodic boundary conditions.
The learning objective is to approximate the solution operator $G$ mapping the initial condition $u(x)$ to the corresponding spatiotemporal solution $s(t,x)$. 

\begin{figure}[H]
  \centering
  \includegraphics[width=0.56\textwidth]{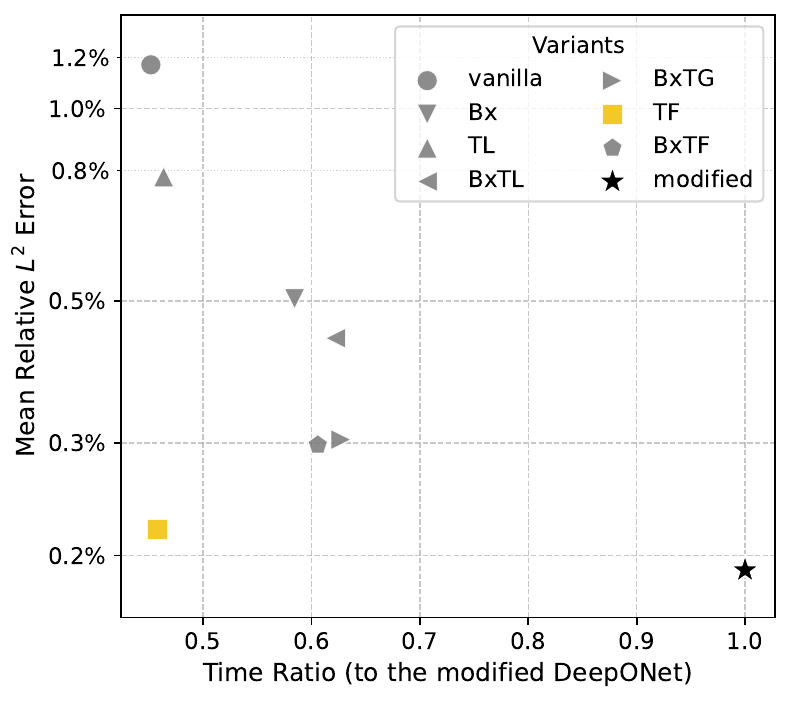}
  \caption{Performance comparison of model variants for the Burgers' equation ($\nu = 10^{-2}$). 
  The horizontal axis shows the average per-iteration training time ratio relative to the modified DeepONet, averaged over multiple random seeds (lower is faster), while the vertical axis shows the mean relative $L^2$ error (lower is better). 
  Marker shapes distinguish different model variants; the modified DeepONet is shown in black, and the best-performing variant is highlighted in gold.}
  \label{fig_burgers_0.01_variants_comparison}
\end{figure}

\begin{figure}[H]
  \centering
  \includegraphics[width=0.52\textwidth]{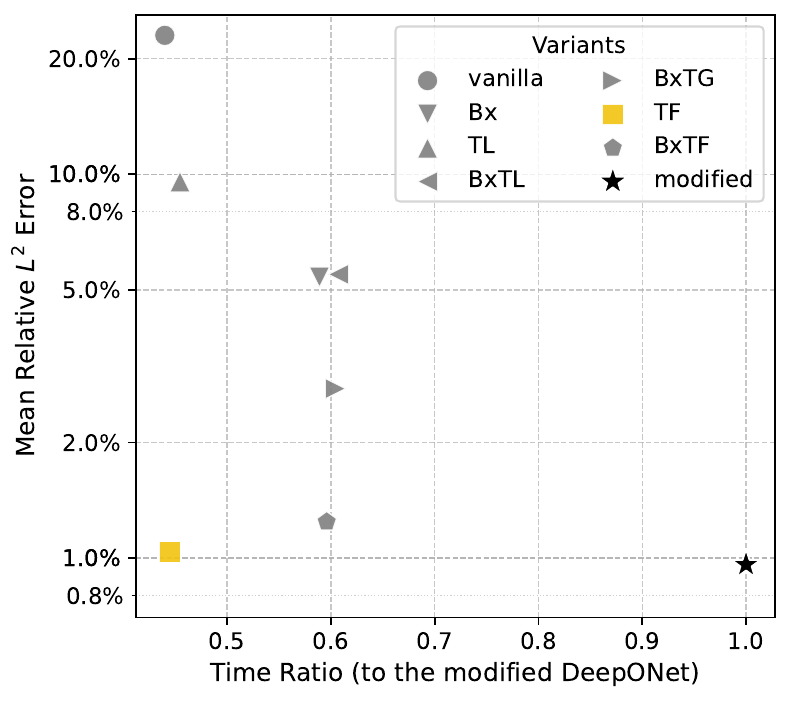}
  \caption{Performance comparison of model variants for the Burgers' equation ($\nu = 10^{-3}$). 
  The horizontal axis shows the average per-iteration training time ratio relative to the modified DeepONet, averaged over multiple random seeds (lower is faster), while the vertical axis shows the mean relative $L^2$ error (lower is better). 
  Marker shapes distinguish different model variants; the modified DeepONet is shown in black, and the best-performing variant is highlighted in gold.}
  \label{fig_burgers_0.001_variants_comparison}
\end{figure}

\begin{figure}[H]
  \centering
  \includegraphics[width=0.52\textwidth]{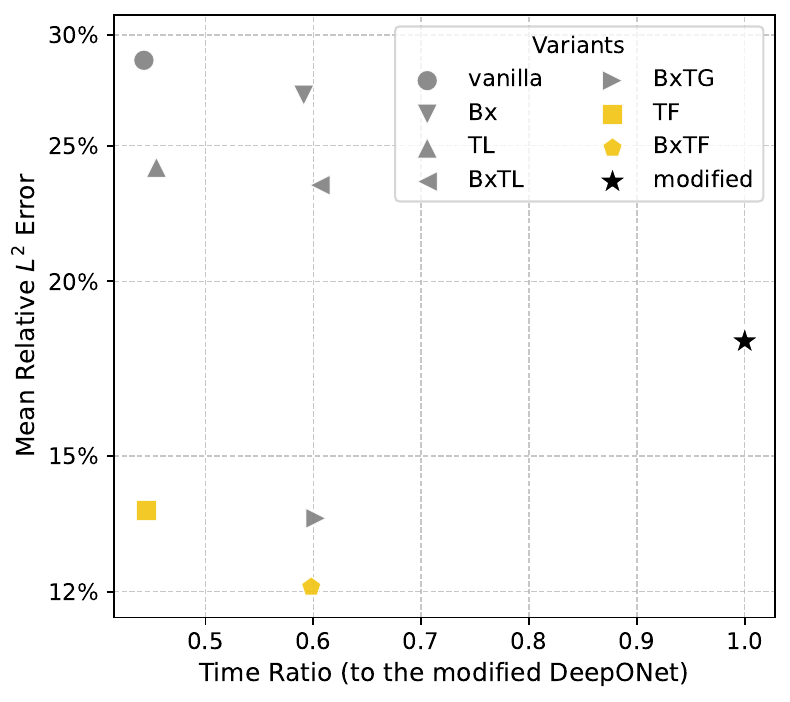}
  \caption{Performance comparison of model variants for the Burgers' equation ($\nu = 10^{-4}$). 
  The horizontal axis shows the average per-iteration training time ratio relative to the modified DeepONet, averaged over multiple random seeds (lower is faster), while the vertical axis shows the mean relative $L^2$ error (lower is better). 
  Marker shapes distinguish different model variants; the modified DeepONet is shown in black, and the best-performing variants are highlighted in gold.}
  \label{fig_burgers_0.0001_variants_comparison}
\end{figure}

\paragraph{Variants comparison.}
\Cref{fig_burgers_0.01_variants_comparison,fig_burgers_0.001_variants_comparison,fig_burgers_0.0001_variants_comparison} 
compare the average per-iteration training time against the mean relative $L^2$ error for the Burgers' equation with viscosities $\nu=10^{-2}$, $10^{-3}$, and $10^{-4}$, respectively. 
Across all three viscosities, the Transformer-inspired models shift markedly to the left of the modified DeepONet, indicating substantial reductions in per-iteration time cost. Again, this speedup can be attributed to the simpler network architectures of the variants.
In terms of accuracy, the Variant TF remains close to the modified DeepONet at $\nu=10^{-2}$ and $\nu=10^{-3}$ (differences within a small margin), and becomes better at $\nu=10^{-4}$. 
Overall, Variant TF offers a favorable speed--accuracy trade-off across viscosities: large gains in training speed while maintaining comparable accuracy in the moderate-/high-viscosity regimes and achieving a modest accuracy improvement in the low-viscosity regime. 
The Variant BxTF also achieves competitive predictive accuracy; at $\nu = 10^{-4}$, it attains the lowest mean relative $L^2$ error of approximately $12\%$, while also maintaining a fast per-iteration training speed.
Both Variants TF and BxTF incorporate the leading Fourier coefficients of the initial condition into the trunk network, effectively introducing a low-pass filter that emphasizes the dominant low-frequency modes and discards high-frequency noise, conceptually similar to Fourier-based neural operator approaches \autocite{Li-2020}.
In contrast, Variant BxTG embeds the entire initial condition into the trunk network, which often yields lower predictive accuracy compared to Variants TF or BxTF. 
For Burgers' equation in the viscosity regimes considered here, the dynamics are primarily governed by the low-frequency Fourier modes of the initial condition, while high-frequency components are rapidly damped by viscosity (even at $\nu=10^{-4}$, nonlinear energy transfer continues to favor lower modes) \autocite{Frisch-2001}. 
Including the full initial condition---containing high-frequency noise---may dilute the emphasis on these dynamically relevant structures, leading to reduced accuracy.

\begin{figure}[H]
  \centering
  \includegraphics[width=0.56\textwidth]{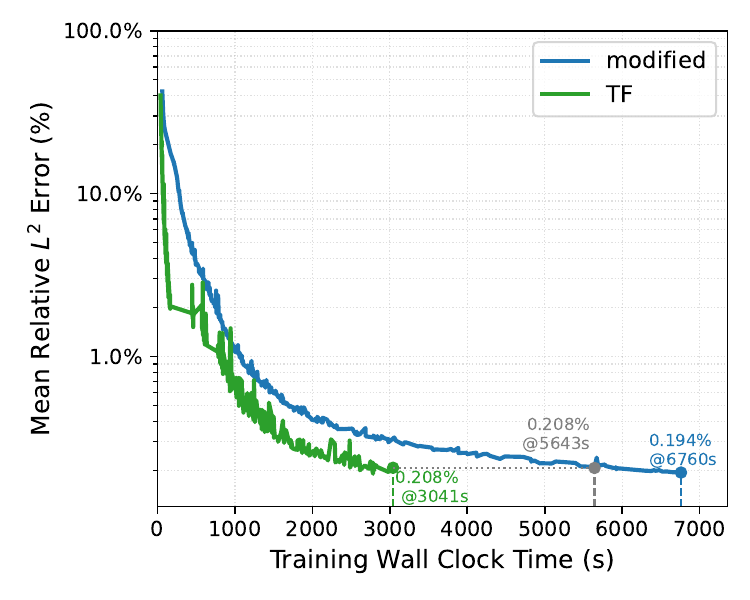}
  \caption{Comparison of training efficiency for the Burgers' equation ($\nu = 10^{-2}$). 
  The horizontal axis shows the wall-clock training time for a single representative run, and the vertical axis shows the mean relative $L^2$ error over all test cases (log scale). 
  Dashed vertical lines and annotations mark the time at which each model completed training iterations. 
  The dotted horizontal line indicates the wall-clock time at which the modified DeepONet reaches the same final accuracy as the TF variant.}
  \label{fig_burgers_0.01_efficiency_comparison}
\end{figure}

\begin{figure}[H]
  \centering
  \includegraphics[width=0.56\textwidth]{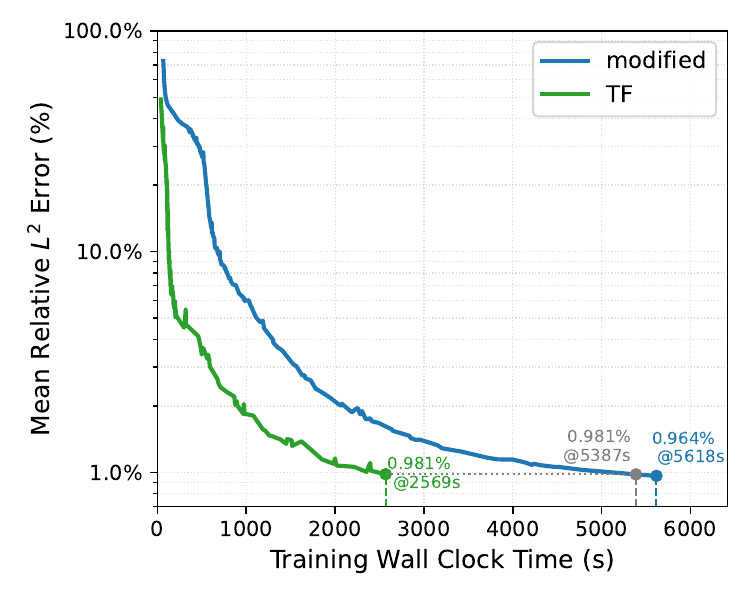}
  \caption{Comparison of training efficiency for the Burgers' equation ($\nu = 10^{-3}$). 
  The horizontal axis shows the wall-clock training time for a single representative run, and the vertical axis shows the mean relative $L^2$ error over all test cases (log scale). 
  Dashed vertical lines and annotations mark the time at which each model completed training iterations. 
  The dotted horizontal line indicates the wall-clock time at which the modified DeepONet reaches the same final accuracy as the TF variant.}
  \label{fig_burgers_0.001_efficiency_comparison}
\end{figure}

\begin{figure}[H]
  \centering
  \includegraphics[width=0.56\textwidth]{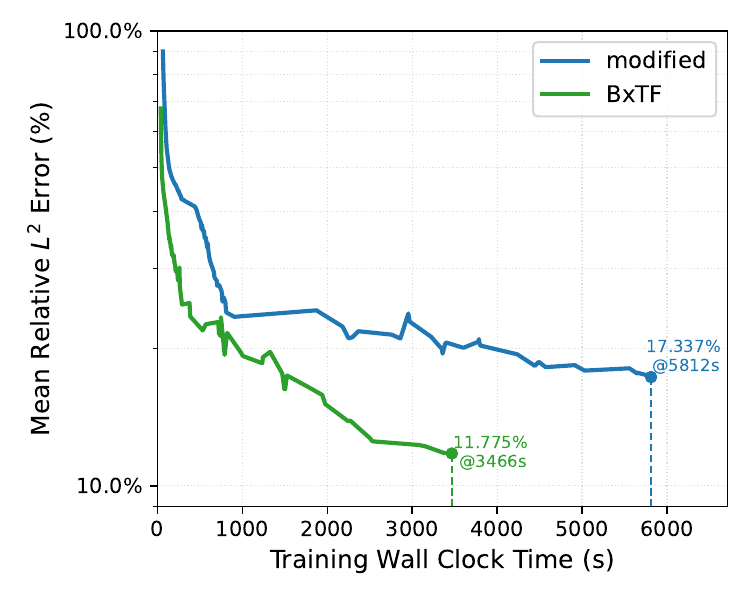}
  \caption{Comparison of training efficiency for the Burgers' equation ($\nu = 10^{-4}$). 
  The horizontal axis shows the wall-clock training time for a single representative run, and the vertical axis shows the mean relative $L^2$ error over all test cases (log scale). 
  Dashed vertical lines and annotations mark the time at which each model completed training iterations.}
  \label{fig_burgers_0.0001_efficiency_comparison}
\end{figure}

\paragraph{Training efficiency.}
\Cref{fig_burgers_0.01_efficiency_comparison,fig_burgers_0.001_efficiency_comparison,fig_burgers_0.0001_efficiency_comparison} 
compare the convergence of the mean relative $L^2$ error with respect to wall-clock training time for selected variants at viscosities $\nu=10^{-2}$, $10^{-3}$, and $10^{-4}$, respectively. 
Across all viscosities, the selected variants consistently reduce mean relative $L^2$ error more rapidly than the modified DeepONet, with all models trained for $200{,}000$ iterations. 
At $\nu=10^{-2}$, the Variant TF attains its final mean relative $L^2$ error of $0.208\%$ in \( 3,041 \,\mathrm{s} \), whereas the modified DeepONet requires \( 5,643 \,\mathrm{s} \) to reach the same accuracy (as indicated by the horizontal dotted line). 
At $\nu=10^{-3}$, Variant TF achieves its final mean relative $L^2$ error of $0.981\%$ in \( 2,569 \,\mathrm{s} \), compared to \( 5,387 \,\mathrm{s} \) for the modified DeepONet to attain the same level. 
At $\nu=10^{-4}$, Variant BxTF reaches a final mean relative $L^2$ error of $11.775\%$ in \( 3,466 \,\mathrm{s} \), while the modified DeepONet converges more slowly to $17.337\%$ in \( 5,812 \,\mathrm{s} \). 
These results underscore that Variants TF and BxTF deliver substantial reductions in training time while maintaining---and, in the low-viscosity regime, even improving---predictive accuracy.

\begin{figure}[H]
  \centering
  \includegraphics[width=0.5\textwidth]{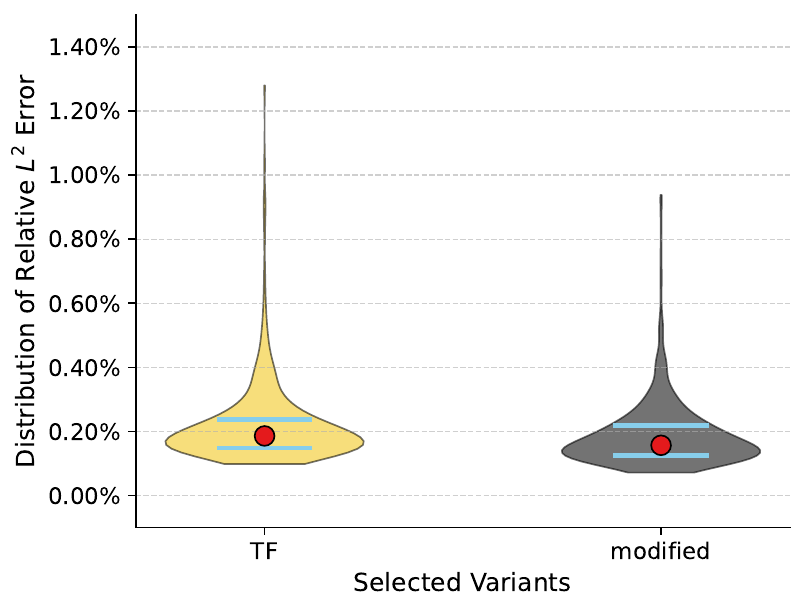}
  \caption{Distribution of relative $L^2$ errors for selected variants (gold: Variant TF; black: the modified DeepONet) on the Burgers' equation ($\nu = 10^{-2}$), evaluated over multiple random seeds and test instances. Horizontal bars span the interquartile range (25th to 75th percentile), and red dots mark the median.}
  \label{fig_burgers_0.01_violin}
\end{figure}

\begin{figure}[H]
  \centering
  \includegraphics[width=0.56\textwidth]{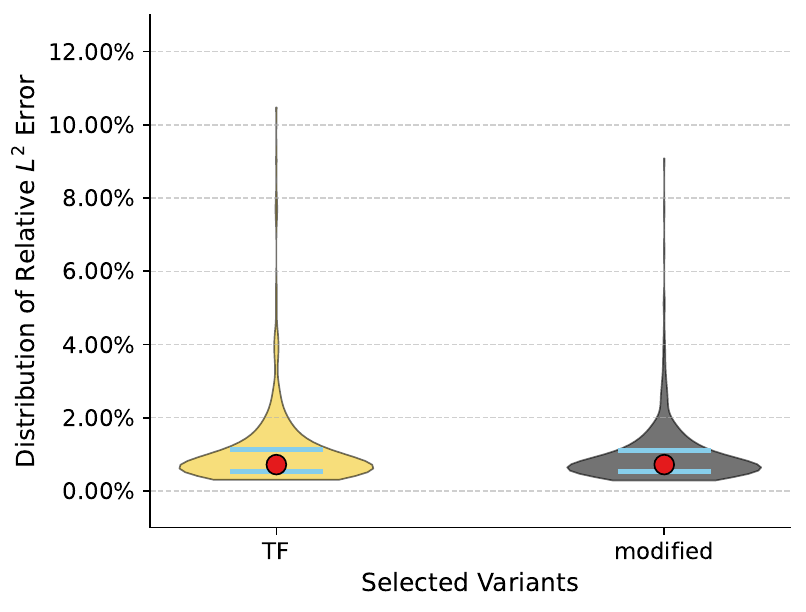}
  \caption{Distribution of relative $L^2$ errors for selected variants (gold: Variant TF; black: the modified DeepONet) on the Burgers' equation ($\nu = 10^{-3}$), evaluated over multiple random seeds and test instances. Horizontal bars span the interquartile range (25th to 75th percentile), and red dots mark the median.}
  \label{fig_burgers_0.001_violin}
\end{figure}

\begin{figure}[H]
  \centering
  \includegraphics[width=0.56\textwidth]{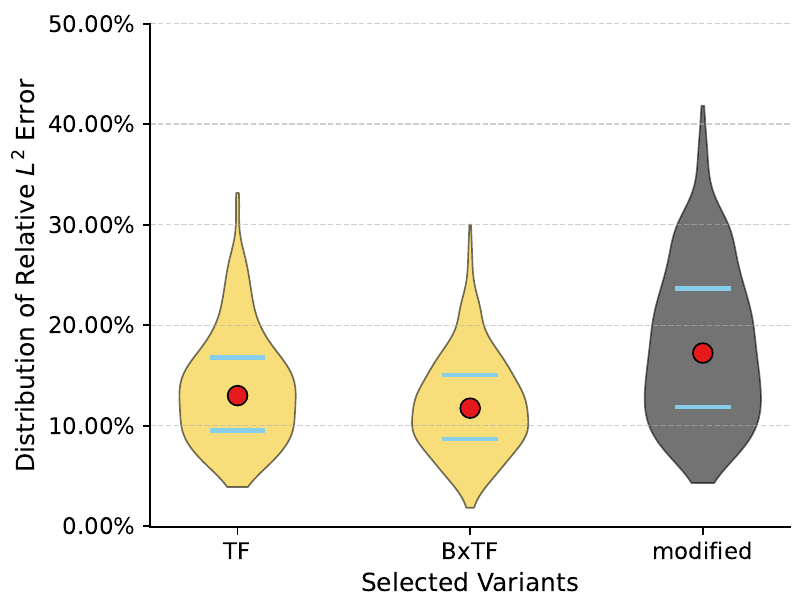}
  \caption{Distribution of relative $L^2$ errors for selected variants (gold: best-performing variants; black: the modified DeepONet) on the Burgers' equation ($\nu = 10^{-4}$), evaluated over multiple random seeds and test instances. Horizontal bars span the interquartile range (25th to 75th percentile), and red dots mark the median.}
  \label{fig_burgers_0.0001_violin}
\end{figure}

\paragraph{Error distribution and equivalence analysis.}
\Cref{fig_burgers_0.01_violin,fig_burgers_0.001_violin,fig_burgers_0.0001_violin} show the distributions of the relative $L^2$ error over all test instances for selected variants at viscosities $\nu=10^{-2}$, $10^{-3}$, and $10^{-4}$, respectively. 
At $\nu=10^{-2}$ (\Cref{fig_burgers_0.01_violin}), the Variant~TF exhibits a median relative $L^2$ error slightly higher than that of the modified DeepONet. 
A non-parametric Wilcoxon two one-sided tests (TOST) (see \Cref{apdx3_data_analysis} for the setup) rejects equivalence between Variant TF and the modified DeepONet with a margin of $\pm 0.01\%$ at this viscosity; the modified DeepONet achieves better accuracy, with a non-parametric Glass's $\Delta = 0.397$, corresponding to a small-to-moderate effect size. 
Thus, the accuracy compromise of Variant TF is modest, especially considering its substantially faster training speed. 
The Spearman correlation is strong ($\rho = 0.858$), indicating similar error patterns between Variant TF and the modified DeepONet.
At $\nu=10^{-3}$ (\Cref{fig_burgers_0.001_violin}), the Variant TF attains a median relative $L^2$ error comparable to that of the modified DeepONet. 
A non-parametric TOST for equivalence, with a margin of $\pm 0.06\%$, confirms that the two models are statistically equivalent in predictive accuracy, while the error patterns remain strongly correlated (Spearman $\rho = 0.869$).
At $\nu=10^{-4}$ (\Cref{fig_burgers_0.0001_violin}), both Variants TF and BxTF achieve lower median relative $L^2$ errors than the modified DeepONet, with overall error distributions shifted downward relative to the modified DeepONet. 
In test cases, Variant BxTF achieved better predictive accuracy in $83.00\%$ of cases compared to the modified DeepONet.
The Wilcoxon TOST rejects equivalence in favor of the variants: Glass's $\Delta = -0.393$ for Variant TF and $\Delta = -0.568$ for Variant BxTF, indicating small-to-moderate improvements in predictive accuracy than the modified DeepONet (negative values indicate the modified DeepONet performs worse). 
The corresponding Spearman correlations are moderate to strong ($\rho = 0.515$ for TF vs.\ modified DeepONet, and $\rho = 0.542$ for BxTF vs.\ modified DeepONet), showing that their error variations remain partially aligned with those of the modified DeepONet across test cases.

\subsection{Korteweg--de Vries Equation}

We finally consider the Korteweg--de Vries (KdV) equation,
\begin{equation}
    \frac{\partial s}{\partial t} + s \frac{\partial s}{\partial x} + \delta^2 \frac{\partial^3 s}{\partial x^3} = 0, 
    \quad (t,x) \in (0, 1)\times(0, 2\pi),
\end{equation}
subject to periodic boundary conditions and the initial condition $s(0,x) = u(x)$. 
The initial conditions are parameterized as $u(x) = a\sin(x) + b\cos(x)$, where $a$ and $b$ are drawn from the standard normal distribution. 
Here, $\delta = 0.1$ denotes the dispersion coefficient. 
The learning objective is to approximate the solution operator $G$ that maps the initial condition $u(x)$ to the corresponding spatiotemporal solution $s(t,x)$.

\begin{figure}[H]
  \centering
  \includegraphics[width=0.56\textwidth]{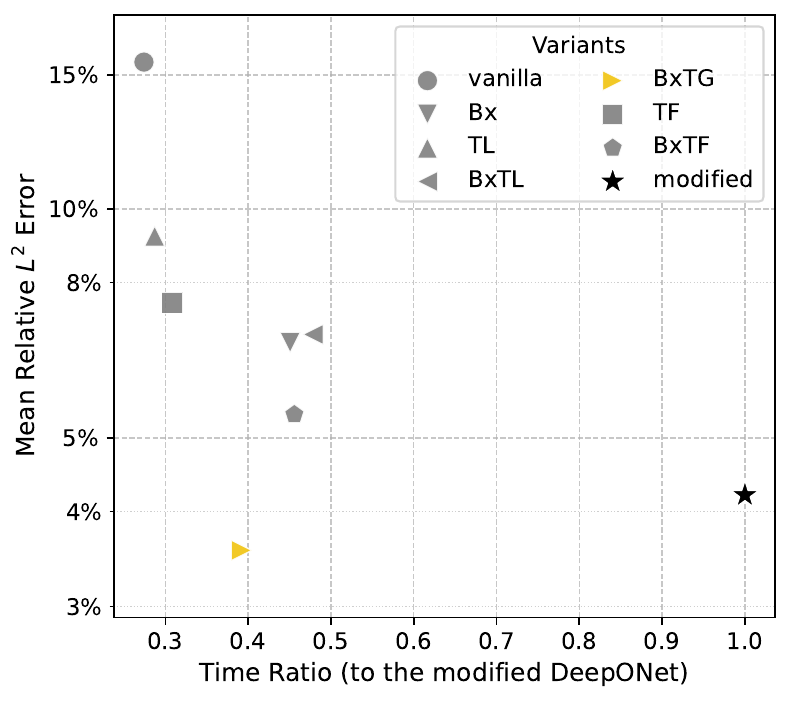}
  \caption{Performance comparison of model variants for the KdV equation. 
  The horizontal axis shows the average per-iteration training time ratio relative to the modified DeepONet, averaged over multiple random seeds (lower is faster); the vertical axis shows the mean relative $L^2$ error (lower is better). 
  Marker shapes distinguish different model variants; the modified DeepONet is shown in black, and the best-performing variant is highlighted in gold.}
  \label{fig_kdv_comparison}
\end{figure}

\begin{figure}[H]
  \centering
  \includegraphics[width=0.56\textwidth]{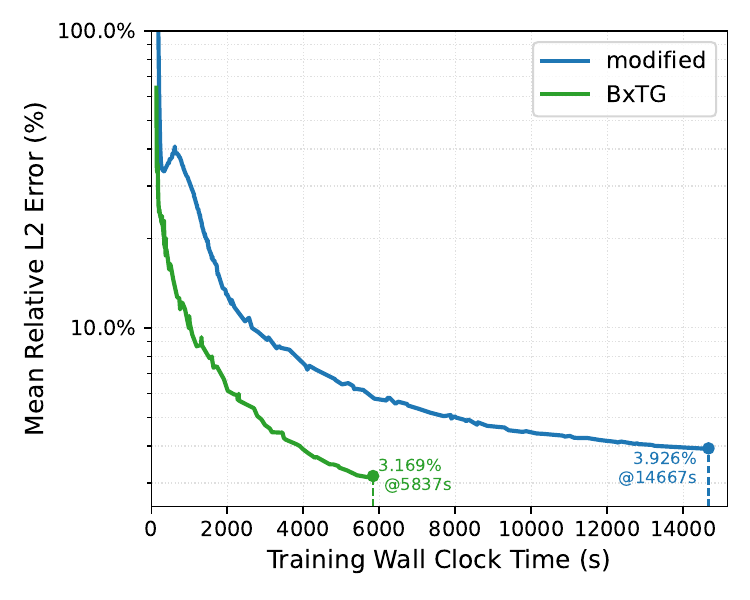}
  \caption{Comparison of training efficiency for the KdV equation. 
  The horizontal axis shows the wall-clock training time for a single representative run, and the vertical axis shows the mean relative $L^2$ error over all test cases (log scale). 
  Dashed vertical lines and annotations mark the reported final point for each run.}
  \label{fig_kdv_timecurve}
\end{figure}

\begin{figure}[H]
  \centering
  \includegraphics[width=0.56\textwidth]{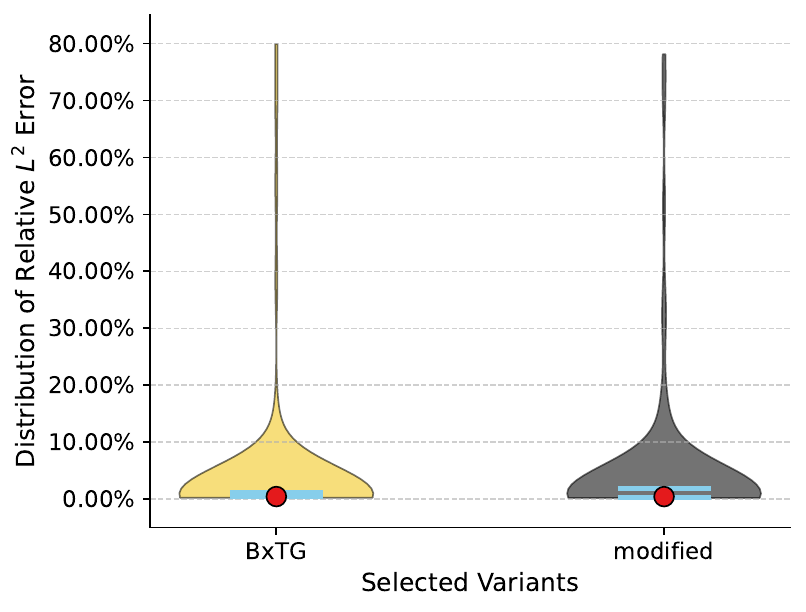}
  \caption{Distribution of relative $L^2$ errors for selected variants (gold: Variant BxTG; black: the modified DeepONet) on the KdV equation, evaluated over multiple random seeds and test instances. 
  Horizontal bars span the interquartile range (25th to 75th percentile), and red dots mark the median.}
  \label{fig_kdv_violin}
\end{figure}

\medskip
The performance of all evaluated models is summarized in \Cref{fig_kdv_comparison,fig_kdv_timecurve,fig_kdv_violin}, with the modified DeepONet (black star marker in~\Cref{fig_kdv_comparison}) serving as the reference for both training time and accuracy. 
The KdV equation is particularly challenging to solve due to the possible presence of soliton solutions. Although in our current setting ($\delta=0.1$ and low-frequency Fourier initial conditions) strong soliton interactions are less pronounced, the KdV equation in general is known to admit soliton solutions, making it challenging to solve accurately.
Among the Transformer-inspired variants, Variant BxTG (gold marker in~\Cref{fig_kdv_comparison}) achieves the lowest mean relative $L^2$ error about $3.55\%$ among the faster models, offering an average per-iteration speedup of about $60\%$ while yielding slightly lower error than the $4.20\%$ achieved by the modified DeepONet. 
The fast training speed of the Transformer-inspired variants is largely attributable to their simpler architectures. 
For the KdV equation, which is highly sensitive to initial conditions, feeding the full initial condition directly into the trunk network---as done in Variant BxTG---likely improves predictive accuracy. In contrast, in our setup, the initial conditions are only first-order Fourier modes, so feeding their coefficients alone (as in Variants TF and BxTF) may not provide sufficiently detailed information for optimal training.
As an illustrative example, the convergence curves in \Cref{fig_kdv_timecurve} show that Variant BxTG reaches a final mean relative $L^2$ error of $3.169\%$ in \( 5,837 \,\mathrm{s} \), compared to $3.926\%$ in \( 14,667 \,\mathrm{s} \) for the modified DeepONet---representing a $60\%$ reduction in wall-clock time, along with a lower final mean relative $L^2$ error. 
Moreover, at any given wall-clock time, the mean relative $L^2$ error of Variant BxTG (averaged over all test instances) remains consistently below that of the modified DeepONet, indicating faster convergence toward the optimum.
Error distributions over all test instances (\Cref{fig_kdv_violin}) show that BxTG attains a similar median relative $L^2$ error to the modified DeepONet ($0.45\%$ and $0.44\%$, respectively), with comparable spread.
From \Cref{fig_kdv_violin}, we observe that both error distributions are concentrated near small values, with low medians and 75th percentiles. 
However, a few large-error cases are present, with relative $L^2$ errors approaching $80\%$. 
These large errors for both models tend to occur in regions where the ground-truth solution exhibits steep gradients; see \Cref{apdx2_raw_data} for details.
A non-parametric Wilcoxon two one-sided test (TOST) with an equivalence margin of $\pm 0.04\%$ concludes that Variant BxTG and the modified DeepONet are \emph{not} equivalent in predictive accuracy: 
Variant BxTG has lower perdictive error. The non-parametric Glass's $\Delta$ is $-0.054$, indicating that Variant BxTG is only a little bit better in accuracy than the modified DeepONet.
In test cases, Variant BxTG achieved better predictive accuracy in $56.00\%$ of cases compared to the modified DeepONet.
Furthermore, a Spearman correlation coefficient of $\rho = 0.860$ indicates a strong positive correlation in the error patterns of the two models.

\paragraph{Overall:} Variant BxTG delivers measurably better predictive accuracy than the modified DeepONet for the KdV equation, while substantially reducing training time.

\section{Discussion and Conclusion}\label{sec_discussion}

We proposed a series of \emph{Transformer-inspired} variants of DeepONet that balance the trade-off between \emph{predictive accuracy} and \emph{computational efficiency} observed in the ``vanilla'' {DeepONet} and the modified DeepONet models.  
By injecting query-point information into the branch network and information of input function into the trunk network, our variants introduce \emph{bidirectional cross-conditioning} between the branch and trunk networks, enabling dynamic dependencies while preserving the simplicity and efficiency of the ``vanilla'' {DeepONet}.  
These modifications are \emph{structurally simple} and \emph{non-intrusive}, requiring only minimal changes to the overall architecture.

Across the four PDE benchmarks, we observed that for each equation, there exists at least one variant that achieves accuracy comparable to (or better than) the modified {DeepONet} while offering improved training efficiency. Notably, the best-performing variant for each PDE appears to be closely related to the characteristics of the underlying equation.  
For example, in the case of Burgers' equation, the most effective variant (TF) injects the leading Fourier coefficients of the initial condition into the trunk network. This design is theoretically motivated: the dynamics of the viscous Burgers' equation are primarily governed by the low-frequency Fourier modes of the initial condition, while high-frequency components are rapidly damped due to viscosity. As a result, making the trunk network explicitly aware of the dominant Fourier modes helps it form more effective query-dependent representations, contributing to the improved predictive accuracy observed in Variant TF.

Looking ahead, one promising direction is to further refine the form of cross-conditioning between the input function and the trunk network. We experimented with injecting single-point evaluations of $u(x)$, the full function values, or a set of Fourier coefficients into the trunk. Inspired by the concept of the domain of dependence in PDEs, future work may explore supplying the trunk network with input-function information restricted to \emph{a local neighborhood} of the query spatial point $x$. Such localized conditioning could yield better alignment with the underlying physics of the equation---particularly for hyperbolic PDEs---while preserving the training efficiency. This perspective may also serve as a bridge between classical numerical methods and neural operator design, offering a principled direction for unifying traditional numerical methods and neural operator design.

\section*{Code and Data Availability}
\addcontentsline{toc}{section}{\protect\numberline {}{Code and Data Availability}}
    To support reproducibility and facilitate future research, the code and all accompanying data will be made publicly available upon publication acceptance.
  
\section*{Acknowledgments}
\addcontentsline{toc}{section}{\protect\numberline {}{Acknowledgement}}

The work of Zhi-Feng Wei is supported by the U.S.\  Department of Energy (DOE) Office of Advanced Scientific Computing Research (ASCR) through the ASCR Distinguished Computational Mathematics Postdoctoral Fellowship (Project No.\ 71268). 

This project was completed with support from the U.S.\  Department of Energy, Advanced Scientific Computing Research program, under the Scalable, Efficient and Accelerated Causal Reasoning Operators, Graphs and Spikes for Earth and Embedded Systems (SEA-CROGS) project (Project No.\ 80278). 

The computational work was performed using the Pacific Northwest National Laboratory (PNNL) Research Computing facilities.

PNNL is a multi-program national laboratory operated for the U.S.\ Department of Energy (DOE) by Battelle Memorial Institute under Contract No. DE-AC05-76RL01830.

\begin{appendices}
\crefalias{section}{appendix}
\crefalias{subsection}{appendix}

\counterwithin{theorem}{section}
\counterwithin{equation}{section}
\counterwithin{figure}{section}
\counterwithin{table}{section}

\renewcommand{\thetheorem}{\thesection\arabic{theorem}}
\renewcommand{\theequation}{\thesection\arabic{equation}}
\renewcommand{\thefigure}{\thesection\arabic{figure}}
\renewcommand{\thetable}{\thesection\arabic{table}}

\section{Data Generation and Network Parameters}\label{apdx1_details}

In this appendix, we present the procedures and parameters used for data generation, neural network configuration, and implementation details, with the aim of facilitating the reproducibility of our results.
To facilitate comparison with prior work, we adopt settings adapted from prior work: for the advection and Burgers' equations we follow \autocite{Wang-2022}; for the diffusion--reaction equation \autocite{Wang-2021}; and for the KdV equation \autocite{Williams-2024}.

\subsection{Data Generation}

The datasets were generated, corresponding to the advection, diffusion--reaction, Burgers', and Korteweg--de Vries (KdV) equations.

For the advection and diffusion--reaction equations, the spatially varying coefficient $u(x)$---interpreted as the advection velocity in the advection equation and as the source term in the diffusion--reaction equation---was sampled from a Gaussian random field (GRF) with correlation length $l = 0.2$ and output scale $1.0$.  
Reference solutions were computed on a uniform grid with $n_t = 101$ time steps and $n_x = 101$ spatial points.  
The advection equation was solved using the Lax--Wendroff finite-difference scheme \autocite{Iserles-2008}; the training set contains $1{,}000$ functions and the test set contains $100$ functions.
The diffusion--reaction equation employed a second-order implicit finite difference method \autocite{Iserles-2008}; the training set contains $10{,}000$ functions and the test set contains $1{,}000$ functions.  

\begin{table}[h!]
\centering
\caption{Data generation parameters for the Advection and Diffusion--Reaction equations.}
\label{tab:data_generation_fd}
\begin{tabular}{lccccccc}
\toprule
Equation  & Train size & Test size & Spatial domain & $l$ & $n_t$ & $n_x$ \\
\midrule
Advection  & $1{,}000$ & 100 &$[0,1]$ & $0.2$ & $101$ & $101$ \\
Diffusion--Reaction & $10{,}000$ & $1{,}000$ &$[0,1]$ & $0.2$ & $101$ & $101$ \\
\bottomrule
\end{tabular}
\end{table}

The viscous Burgers' equation is considered on the periodic domain $[0,1]$ with viscosities $\nu = 10^{-2}$, $10^{-3}$, and $10^{-4}$. Initial conditions $u(x)$ for Burgers' equation are sampled from a Gaussian random field (GRF) as
\begin{displaymath}
    u \sim \mathcal{N}\bigl(0,\, 25^2(-\Delta+5^2 I)^{-4}\bigr).
\end{displaymath}
The KdV equation is considered on the periodic domain $[0, 2\pi]$ with dispersion parameter $\delta = 0.1$.  
Initial conditions for the KdV equation are parameterized as $u(x) = a \sin x + b \cos x$, where $a$ and $b$ are independent standard normal random variables.  
For both the Burgers' and KdV equations, reference solutions are computed using a spectral Fourier discretization and a fourth-order stiff time-stepping scheme \autocite{Cox-2002}, with a spectral grid of $s = 4{,}096$ points and a time step of $10^{-4}$.  
Solutions are stored on a coarse grid with $n_t$ time levels and $n_x$ spatial points, as listed in \Cref{tab:data_generation_spectral}.  
For Burgers' equation, a total of $1{,}500$ realizations are generated for each viscosity, with $1{,}000$ used for training and $500$ for testing.  
For the KdV equation, a total of $600$ realizations are produced, with $500$ used for training and $100$ for testing.
\begin{table}[h!]
\centering
\caption{Data generation parameters for the Burgers' and KdV equations.}
\label{tab:data_generation_spectral}
\begin{tabular}{lccccccc}
\toprule
Equation & Train size & Test size & Domain & Spectral $s$ & $\Delta t$ & $n_t$ & $n_x$\\
\midrule
Burgers' &1{,}000 & 500 & $[0,1]$ & 4{,}096 & $10^{-4}$ & 101 & 101\\
KdV & 500 & 100 & $[0,2\pi]$ & 4{,}096 & $10^{-4}$ & 101 & 129 \\
\bottomrule
\end{tabular}
\end{table}

\subsection{Network Parameters}
This subsection summarizes the hyperparameters used across all experiments. For clarity, we group them into three categories: (1) training setup in \Cref{tab:train_setup}, (2) model specifications in \Cref{tab:model_specs}, (3) sampling and loss weighting in \Cref{tab:sampling_loss}.

Across all experiments we use the AdamW optimizer, \texttt{tanh} activations, and 64-bit precision. 
To address the challenge of loss balancing in physics-informed training, we employed both the conjugate kernel (CK) weighting strategy with moderate localization (see \autocite{Qadeer-2023,Wang-2022}) and the balanced-residual-decay-rate (BRDR) self-adaptive weighting scheme \autocite{Chen-2025} in our numerical experiments. For each experiment, we report the results obtained from the weighting method that yielded better performance.
All models are implemented and trained using the JAX framework (version~0.5.3), and optimized using Optax (version~0.2.4).

\textit{Fourier feature embeddings.}
For the diffusion--reaction equation, we use random Fourier features with 150 frequency draws; for each draw we include a paired sine--cosine term sharing the same random phase, yielding 300 features in total.
For Burgers' equation viscosities $10^{-2}$, $10^{-3}$, and $10^{-4}$, we use deterministic Fourier embeddings built from paired sine--cosine harmonics up to orders $4$, $6$, and $8$, respectively (inclusive).
For the KdV equation, we use a deterministic Fourier embedding with harmonics up to order $12$ (inclusive).
Therefore, for Burgers' and KdV equations, periodic boundary conditions are satisfied implicitly via the Fourier representation, so no explicit boundary sampling is required.
\begin{table}[H]
\centering
\caption{Training setup.}
\label{tab:train_setup}
\begin{tabular}{lcccc}
\toprule
Parameter & Advection & Diffusion--Reaction & Burgers' & KdV \\
\midrule
Train size        & 1{,}000  & 10{,}000 & 1{,}000 & 500 \\
Test size         & 100      & 1{,}000  & 500     & 100 \\
Iterations        & 300{,}000 & 120{,}000 & 200{,}000 & 200{,}000 \\
Batch size        & 10{,}000 & 10{,}000 & 10{,}000 & 16{,}384 \\
Learning rate     & 0.001    & 0.001    & 0.001    & 0.001 \\
LR schedule       & Exp(500, 0.99)$^{\dagger}$ & Exp(500, 0.99) & Exp(500, 0.99) & Exp(500, 0.99) \\
Optimizer         & AdamW    & AdamW    & AdamW    & AdamW \\
Activation        & \texttt{tanh} & \texttt{tanh} & \texttt{tanh} & \texttt{tanh} \\
Precision (bits)  & 64       & 64       & 64       & 64 \\
\bottomrule
\end{tabular}
\begin{flushleft} 
$^{\dagger}$ Exponential decay with transition steps $500$ and decay rate $0.99$:
$\eta_t=\eta_0\,0.99^{\,t/500}$.
\end{flushleft} 
\end{table}

\begin{table}[H]
\centering
\caption{Model specifications.}
\label{tab:model_specs}
\begin{tabular}{lcccc}
\toprule
Parameter & Advection & Diffusion--Reaction & Burgers' & KdV \\
\midrule
Branch width            & 100 & 50  & 100 & 128 \\
Branch depth            & 6+1$^{\ddagger}$ & 4+1 & 6+1 & 6+1 \\
Trunk width             & 100 & 50  & 100 & 128 \\
Trunk depth             & 6+1 & 4+1 & 6+1 & 6+1 \\
\bottomrule
\end{tabular}
\begin{flushleft} 
$^{\ddagger}$ ``6 hidden +1'' denotes 6 hidden layers plus an output layer.
\end{flushleft} 
\end{table}

\begin{table}[H]
\centering
\caption{Sampling and loss weighting.}
\label{tab:sampling_loss}
\begin{tabular}{lcccc}
\toprule
Parameter & Advection & Diffusion--Reaction & Burgers' & KdV \\
\midrule
Initial allocation points   & 101 & 101 & 101 & 257 \\
Boundary allocation points  & 101 & 101 & NA$^\ast$ & NA$^\ast$ \\
Residual points             & 2{,}500 & 100 & 2{,}500 & $101\times 257$ \\
Weighting scheme            & BRDR  & CK & CK  & CK \\
\bottomrule
\end{tabular}
\begin{flushleft} 
$^\ast$ boundary conditions are enforced implicitly by deterministic Fourier feature embeddings, so no explicit boundary sampling is required.
\end{flushleft} 
\end{table}

\section{Supporting Data and Extended Results}\label{apdx2_raw_data}

\subsection{Raw Data Underlying Figures}\label{apdx21_raw_data}
We provide here the raw numerical data underlying 
\Cref{fig_advection_comparison,fig_dre_comparison,fig_burgers_0.01_variants_comparison,fig_burgers_0.001_variants_comparison,fig_burgers_0.0001_variants_comparison,fig_kdv_comparison}. 
To produce these plots, we compute the mean relative $L^2$ error over all test cases, averaged across all random seeds. 
In addition to the mean relative $L^2$ error, we also report the corresponding standard deviations, which are not reflected in the scatter plots such as \Cref{fig_kdv_comparison}. 
The tables below also list the average training time per iteration (in seconds), which determines the values shown on the horizontal axes of scatter plots.

\begin{table}[H]
  \centering
  \caption{Raw data for the Advection equation. Mean relative $L^2$ error and standard deviation (STD) are computed over all test cases and averaged across all random seeds. Average time per iteration and the ratio relative to the modified DeepONet are also reported.
  }
  \label{tab:advection_raw}
  \begin{tabular}{lccccc}
    \toprule
    Variant & Mean Rel.\ $L^2$ Error (\%) & STD (\%) & Avg.\ Time per Iter.\ (s) & Time Ratio\\
    \midrule
    vanilla   & 2.096 & 0.569 & 0.0089 & 0.492 \\
    Bx        & 1.089 & 0.256 & 0.0097 & 0.534 \\
    TL        & 1.513 & 0.409 & 0.0092 & 0.508 \\
    BxTL      & 1.002 & 0.244 & 0.0100 & 0.552 \\
    BxTG      & 0.925 & 0.322 & 0.0101 & 0.559 \\
    modified  & 0.926 & 0.315 & 0.0181 & 1.000 \\
    \bottomrule
  \end{tabular}
\end{table}

\begin{table}[H]
\centering
\caption{Raw data for the Diffusion--Reaction equation. Mean relative $L^2$ error and standard deviation (STD) are computed over all test cases and averaged across all random seeds. Average time per iteration and the ratio relative to the modified DeepONet are also reported.}
\label{tab:dre_rawdata}
\begin{tabular}{lcccc}
\toprule
Variant & Mean Rel.\ $L^2$ Error (\%) & STD (\%) & Avg.\ Time per Iter.\ (s) & Time Ratio \\
\midrule
vanilla   & 0.224 & 0.097 & 0.0096 & 0.560 \\
Bx        & 0.231 & 0.115 & 0.0118 & 0.690 \\
TL        & 0.190 & 0.091 & 0.0099 & 0.581 \\
BxTL      & 0.262 & 0.133 & 0.0136 & 0.794 \\
BxTG      & 0.277 & 0.134 & 0.0116 & 0.677 \\
modified  & 0.205 & 0.099 & 0.0171 & 1.000 \\
\bottomrule
\end{tabular}
\end{table}

\begin{table}[H]
\centering
\caption{Raw data for the Burgers' equation with viscosity $\nu = 10^{-2}$. 
Mean relative $L^2$ error and standard deviation (STD) are computed over all test cases and averaged across all random seeds. 
Average time per iteration and the ratio relative to the modified DeepONet are also reported.}
\label{tab:burgers_0.01_raw_data}
\begin{tabular}{lcccc}
\toprule
Variant & Mean Rel.\ $L^2$ Error (\%) & STD (\%) & Avg.\ Time per Iter.\ (s) & Time Ratio \\
\midrule
vanilla  & 1.170 & 1.123 & 0.0140 & 0.452 \\
Bx       & 0.504 & 0.358 & 0.0182 & 0.584 \\
TL       & 0.782 & 0.571 & 0.0144 & 0.464 \\
BxTL     & 0.437 & 0.329 & 0.0193 & 0.622 \\
BxTG     & 0.304 & 0.187 & 0.0195 & 0.627 \\
TF       & 0.220 & 0.137 & 0.0142 & 0.458 \\
BxTF     & 0.298 & 0.181 & 0.0188 & 0.606 \\
modified & 0.190 & 0.116 & 0.0311 & 1.000 \\
\bottomrule
\end{tabular}
\end{table}

\begin{table}[H]
\centering
\caption{Raw data for the Burgers' equation with viscosity $\nu = 10^{-3}$. 
Mean relative $L^2$ error and standard deviation (STD) are computed over all test cases and averaged across all random seeds. 
Average time per iteration and the ratio relative to the modified DeepONet are also reported.}
\label{tab:burgers_0.001_raw_data}
\begin{tabular}{lcccc}
\toprule
Variant & Mean Rel.\ $L^2$ Error (\%) & STD (\%) & Avg.\ Time per Iter.\ (s) & Time Ratio \\
\midrule
vanilla  & 23.012 & 14.192 & 0.0131 & 0.440 \\
Bx       &  5.392 &  3.815 & 0.0176 & 0.589 \\
TL       &  9.566 &  7.151 & 0.0136 & 0.455 \\
BxTL     &  5.486 &  4.748 & 0.0181 & 0.607 \\
BxTG     &  2.764 &  2.507 & 0.0181 & 0.604 \\
TF       &  1.038 &  1.116 & 0.0133 & 0.445 \\
BxTF     &  1.244 &  1.187 & 0.0178 & 0.596 \\
modified &  0.962 &  0.920 & 0.0299 & 1.000 \\
\bottomrule
\end{tabular}
\end{table}

\begin{table}[H]
\centering
\caption{Raw data for the Burgers' equation with viscosity $\nu = 10^{-4}$. 
Mean relative $L^2$ error and standard deviation (STD) are computed over all test cases and averaged across all random seeds. 
Average time per iteration and the ratio relative to the modified DeepONet are also reported.}
\label{tab:burgers_0.0001_raw_data}
\begin{tabular}{lcccc}
\toprule
Variant & Mean Rel.\ $L^2$ Error (\%) & STD (\%) & Avg.\ Time per Iter.\ (s) & Time Ratio \\
\midrule
vanilla  & 28.773 & 12.005 & 0.0133 & 0.443 \\
Bx       & 27.163 & 10.892 & 0.0178 & 0.591 \\
TL       & 24.128 &  8.487 & 0.0137 & 0.455 \\
BxTL     & 23.427 & 11.986 & 0.0182 & 0.606 \\
BxTG     & 13.541 &  6.654 & 0.0181 & 0.602 \\
TF       & 13.715 &  6.805 & 0.0134 & 0.445 \\
BxTF     & 12.094 &  6.060 & 0.0180 & 0.598 \\
modified & 18.119 & 10.371 & 0.0301 & 1.000 \\
\bottomrule
\end{tabular}
\end{table}

\begin{table}[H]
\centering
\caption{Raw data for the KdV equation. 
Mean relative $L^2$ error and standard deviation (STD) are computed over all test cases and averaged across all random seeds. 
Average time per iteration and the ratio relative to the modified DeepONet are also reported.}
\label{tab:kdv_raw_data}
\begin{tabular}{lcccc}
\toprule
Variant & Mean Rel.\ $L^2$ Error (\%) & STD (\%) & Avg.\ Time per Iter.\ (s) & Time Ratio \\
\midrule
vanilla  & 15.606 & 17.957 & 0.0223 & 0.274 \\
Bx       &  6.671 & 16.192 & 0.0367 & 0.451 \\
TL       &  9.216 & 15.194 & 0.0234 & 0.287 \\
BxTL     &  6.837 & 16.372 & 0.0389 & 0.478 \\
BxTG     &  3.554 & 12.292 & 0.0319 & 0.392 \\
TF       &  7.525 & 17.463 & 0.0251 & 0.308 \\
BxTF     &  5.367 & 14.556 & 0.0371 & 0.456 \\
modified &  4.203 & 12.405 & 0.0814 & 1.000 \\
\bottomrule
\end{tabular}
\end{table}

The raw numerical data underlying \Cref{fig_kdv_timecurve,fig_kdv_violin} and other similar figures (e.g., additional efficiency-accuracy curves and full per-instance error distributions) involve extensive time series and individual error records, and are therefore omitted here for brevity. For violin plots, we report in \Cref{tab_violin_summary} the median and interquartile range (Q1, Q3) of the relative $L^2$ error. The full datasets are available from the authors upon request.

\begin{table}[H]
\centering
\caption{Summary statistics (Q1, Median, Q3) of the relative $L^2$ error for selected variants in violin plots. All values are reported in percent.}
\label{tab_violin_summary}
\begin{tabular}{lcccc}
\toprule
Equation & Variant & Q1 (\%) & Median (\%) & Q3 (\%) \\
\midrule
Advection & BxTG & 0.68 & 0.85 & 1.08 \\
          & modified & 0.67 & 0.88 & 1.10 \\
\midrule
Diffusion--Reaction & TL & 0.14 & 0.17 & 0.22 \\
    & modified & 0.15 & 0.18 & 0.24 \\
\midrule
Burgers' ($\nu=10^{-2}$) & TF & 0.15 & 0.19 & 0.24 \\
                        & modified & 0.13 & 0.16 & 0.22 \\
Burgers' ($\nu=10^{-3}$) & TF & 0.54 & 0.72 & 1.13 \\
                        & modified & 0.54 & 0.72 & 1.10 \\
Burgers' ($\nu=10^{-4}$) & TF & 9.52 & 13.00 & 16.77 \\
                        & BxTF & 8.70 & 11.75 & 15.04 \\
                        & modified & 11.89 & 17.2300 & 23.64 \\
\midrule
KdV & BxTG & 0.35 & 0.45 & 1.16 \\
    & modified & 0.31 & 0.44 & 1.87 \\
\bottomrule
\end{tabular}
\end{table} 

\subsection{Reference-Prediction Comparisons and Discussion}\label{apdx22_pred_comparison}
In this section, we visualize representative cases by comparing the reference solution with predictions from selected model variants. For each PDE, we report the modified DeepONet and a top-performing Transformer-inspired variant; each panel shows the reference solution, the model prediction, and the pointwise absolute error.

\begin{figure}[H]
  \centering
  \includegraphics[width=\textwidth]{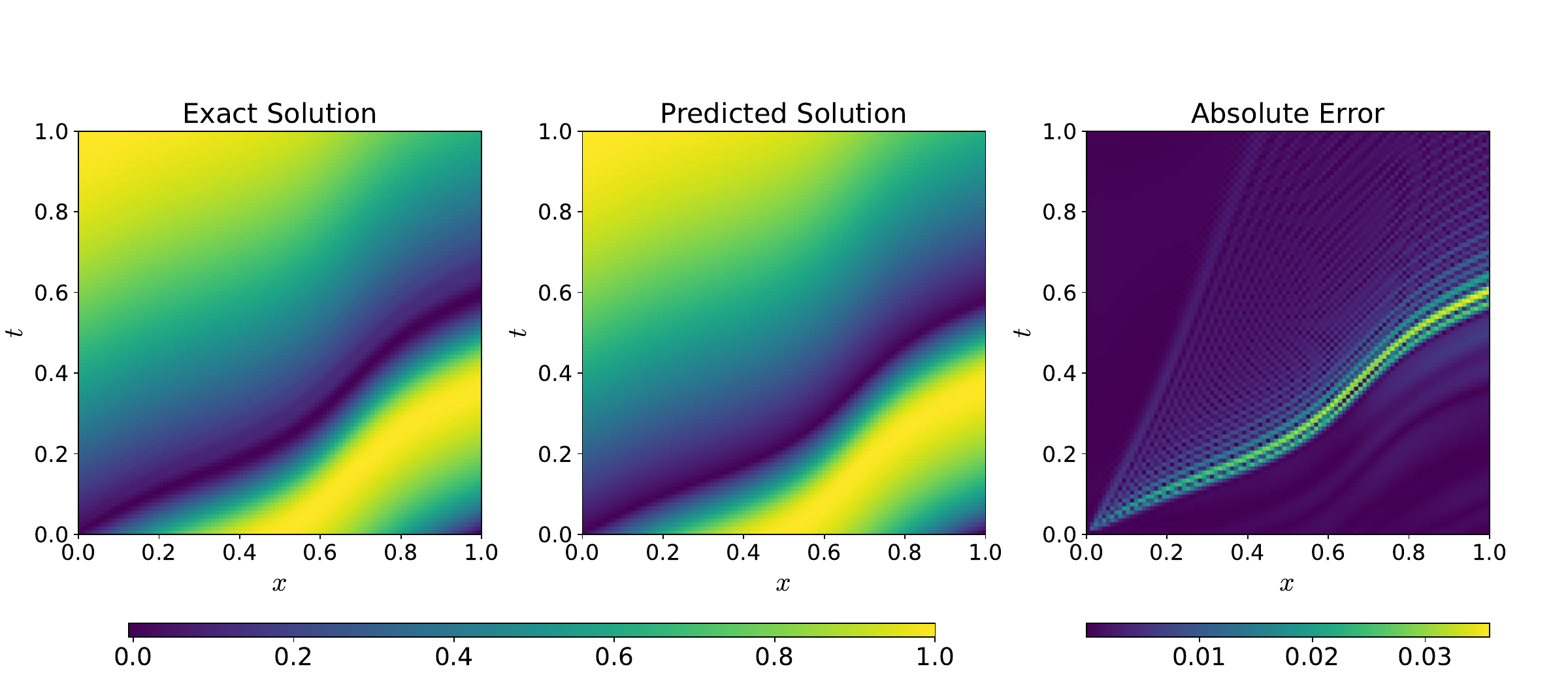}
  \caption{Advection (the modified DeepONet). Left to right: reference solution, model prediction, and absolute error on a representative training instance. The $L^2$ relative error is $0.87\%$.}
  \label{fig:adv_deeponet}
\end{figure}

\begin{figure}[H]
  \centering
  \includegraphics[width=\textwidth]{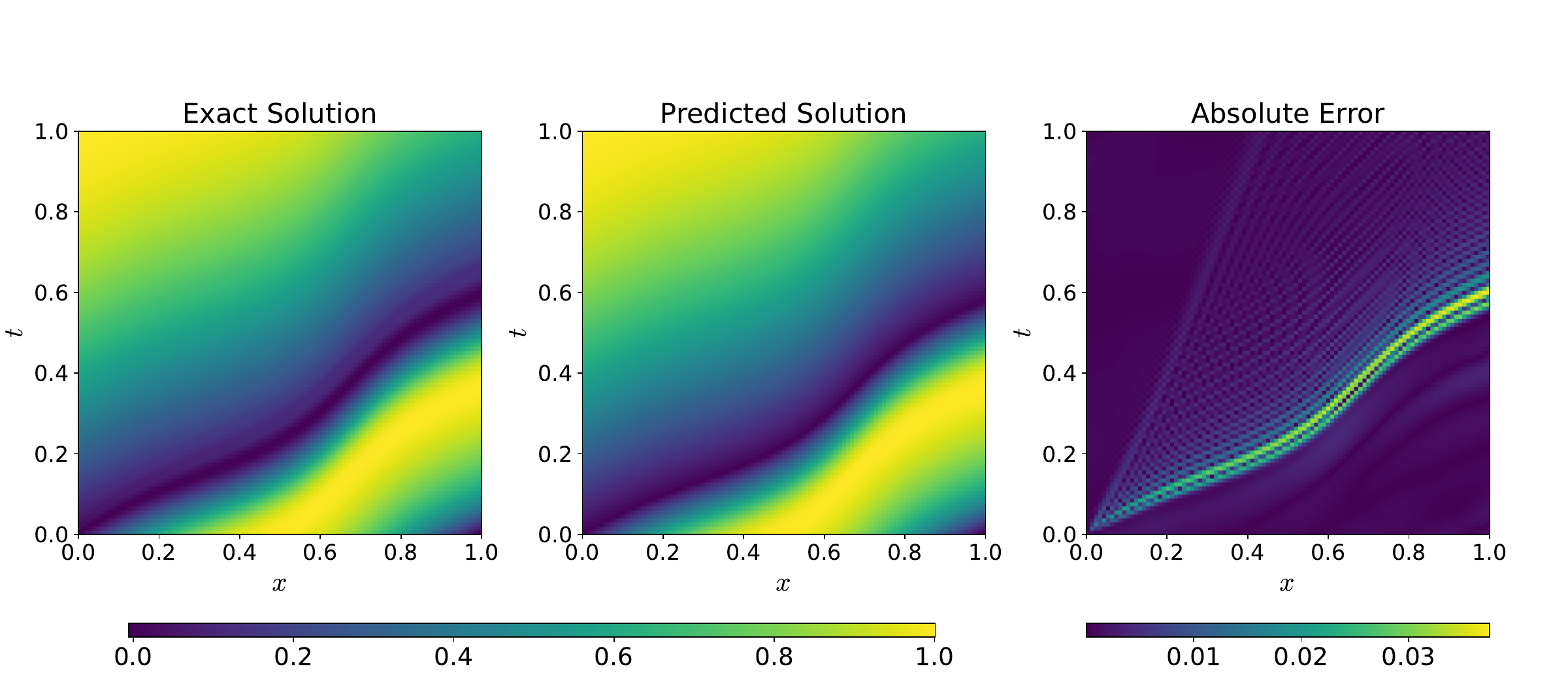}
  \caption{Advection (Variant BxTG). Left to right: reference solution, model prediction, and absolute error on the same instance as in \Cref{fig:adv_deeponet}. The $L^2$ relative error is $0.91\%$.}
  \label{fig:adv_bxtg}
\end{figure}

For the Advection equation, we compare the modified DeepONet with the BxTG variant; see \Cref{fig:adv_deeponet,fig:adv_bxtg}. On the shown instance, the two models achieve comparable accuracy (relative $L^2$ errors $0.87\%$ for the modified DeepONet and $0.91\%$ for the Variant BxTG).

\begin{figure}[H]
  \centering
  \includegraphics[width=\textwidth]{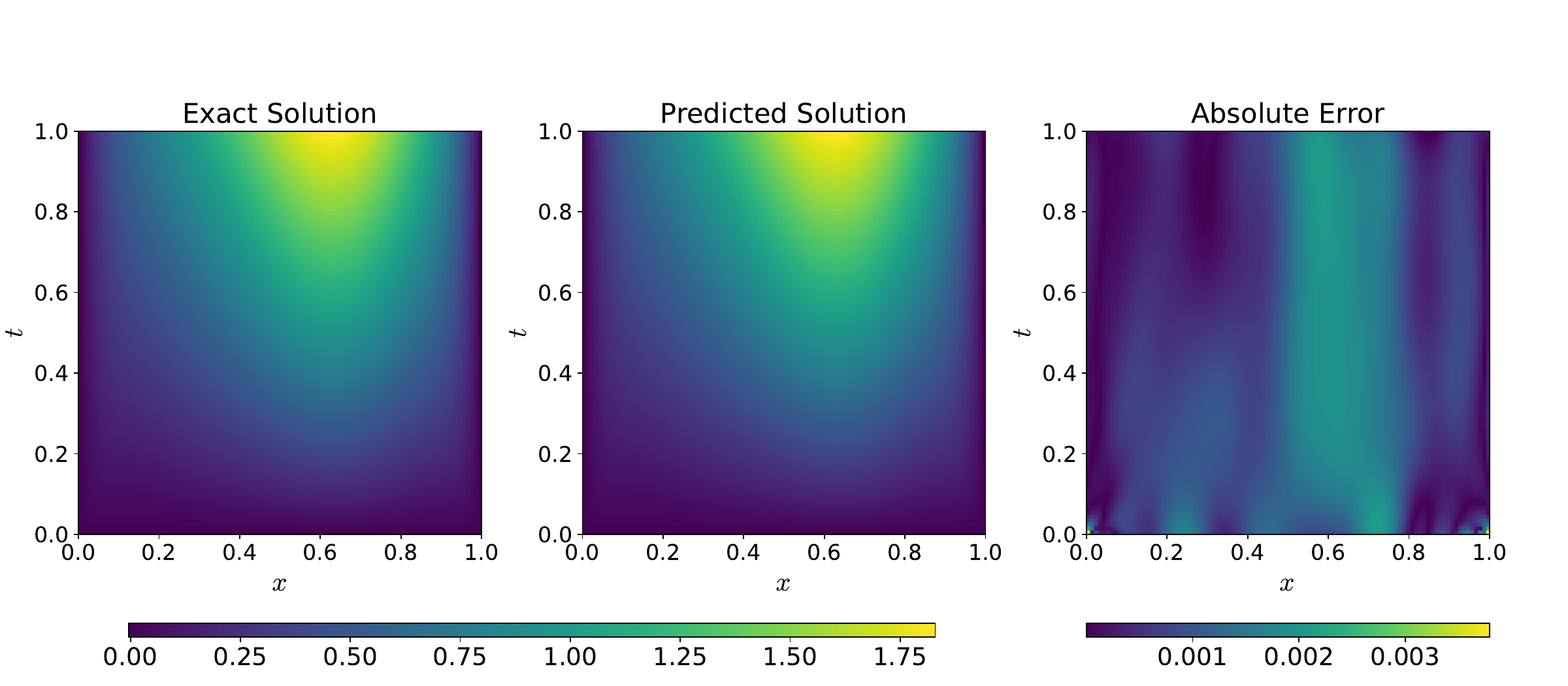}
  \caption{Diffusion--Reaction (the modified DeepONet). Left to right: reference solution, model prediction, and absolute error on a representative training instance. The $L^2$ relative error is $0.14\%$.}
  \label{fig:dre_deeponet}
\end{figure}

\begin{figure}[H]
  \centering
  \includegraphics[width=\textwidth]{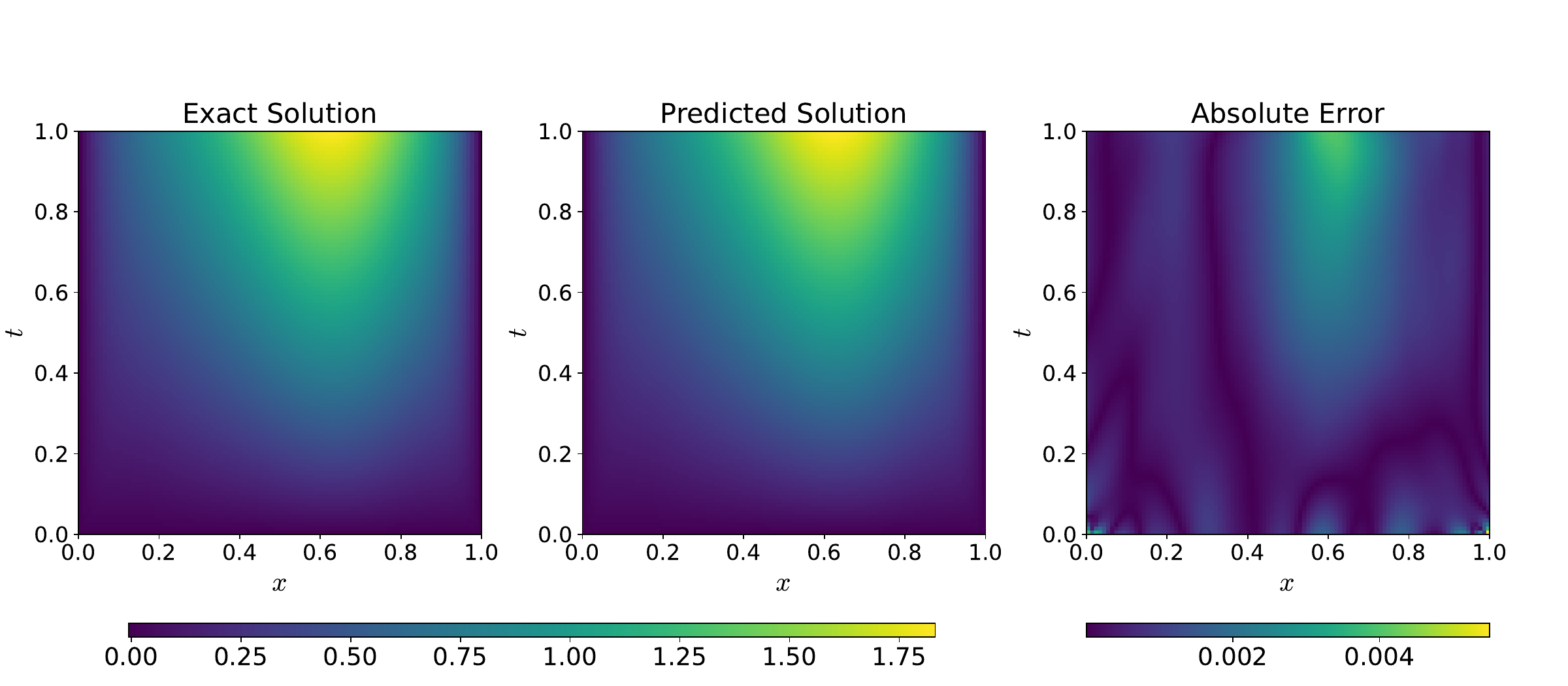}
  \caption{Diffusion--Reaction (Variant TL). Left to right: reference solution, model prediction, and absolute error on the same instance as in \Cref{fig:dre_deeponet}. The $L^2$ relative error is $0.15\%$.}
  \label{fig:dre_tl}
\end{figure}

\clearpage
For the Diffusion--Reaction equation, we compare the modified DeepONet with the TL variant; see \Cref{fig:dre_deeponet,fig:dre_tl}. The two models again attain comparable accuracy (relative $L^2$ errors $0.14\%$ for the modified DeepONet and $0.15\%$ for the Variant TL).

\begin{figure}[H]
  \centering
  \includegraphics[width=\textwidth]{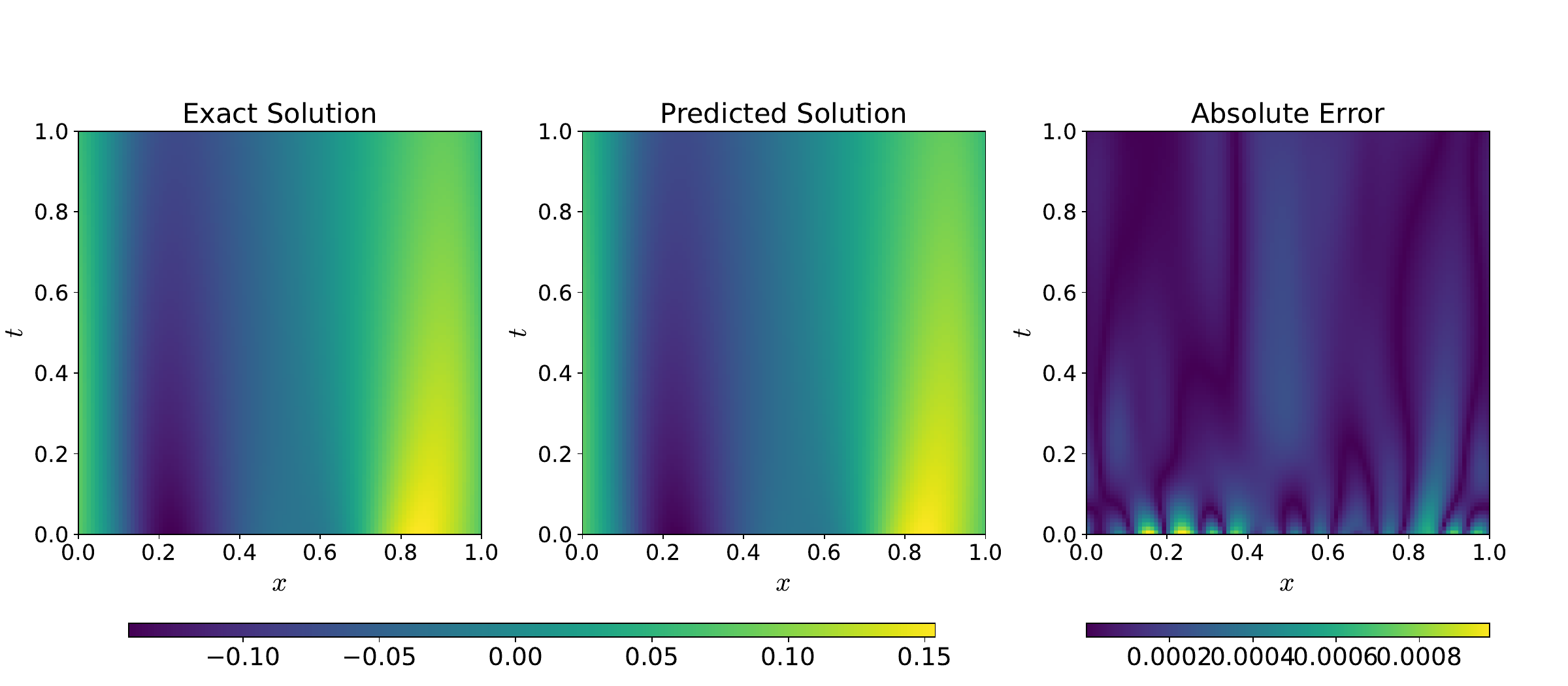}
  \caption{Burgers' equation with viscosity $10^{-2}$ (the modified DeepONet). Left to right: reference solution, model prediction, and absolute error on a representative training instance. The $L^2$ relative error is $0.19\%$.}
  \label{fig:burgers2_deeponet}
\end{figure}

\begin{figure}[H]
  \centering
  \includegraphics[width=\textwidth]{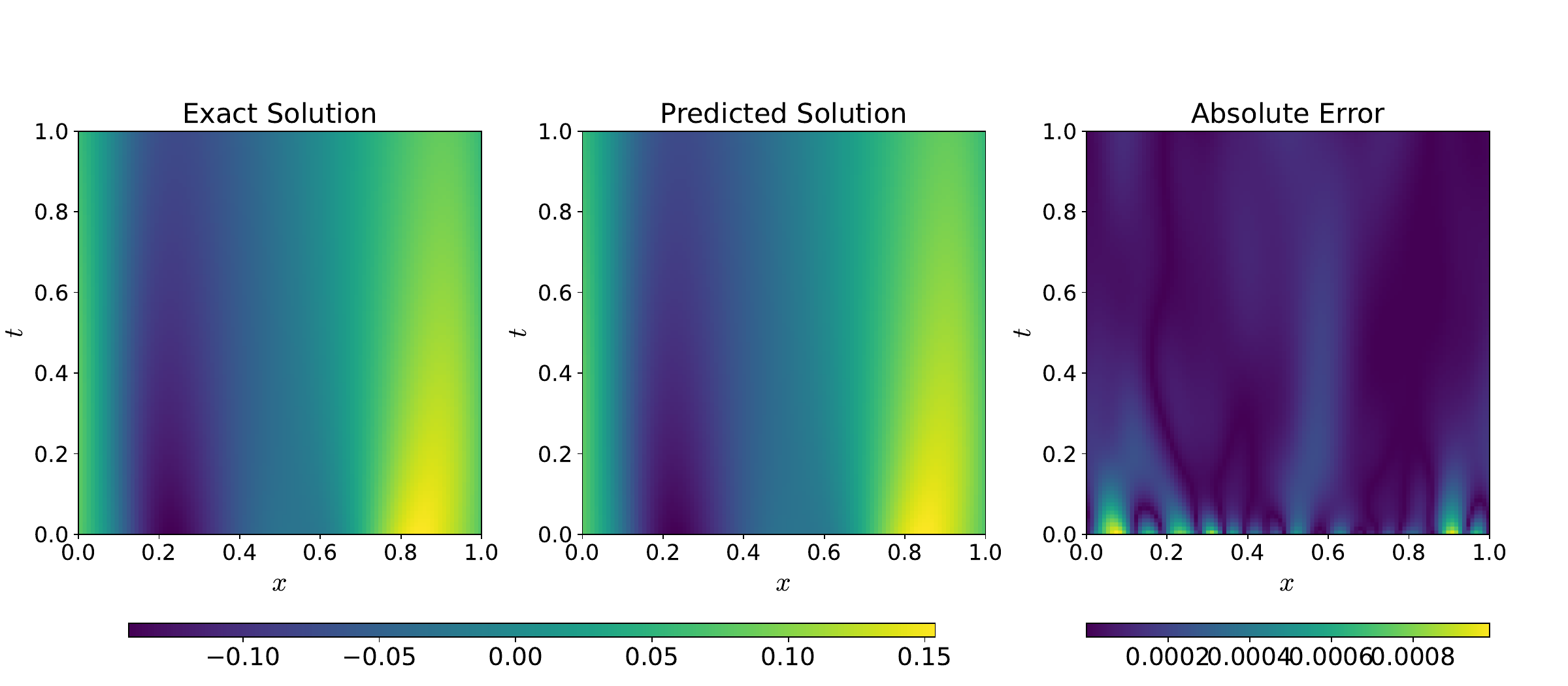}
  \caption{Burgers' equation with viscosity $10^{-2}$ (Variant TF). Left to right: reference solution, model prediction, and absolute error on the same instance as in \Cref{fig:burgers2_deeponet}. The $L^2$ relative error is $0.17\%$.}
  \label{fig:burgers2_tf}
\end{figure}

\clearpage
For Burgers' equation with viscosity $10^{-2}$, we compare the modified DeepONet with Variant TF; see \Cref{fig:burgers2_deeponet,fig:burgers2_tf}. The two methods perform similarly on this instance (relative $L^2$ errors $0.19\%$ for the modified DeepONet and $0.17\%$ for the variant TF).

\begin{figure}[H]
  \centering
  \includegraphics[width=\textwidth]{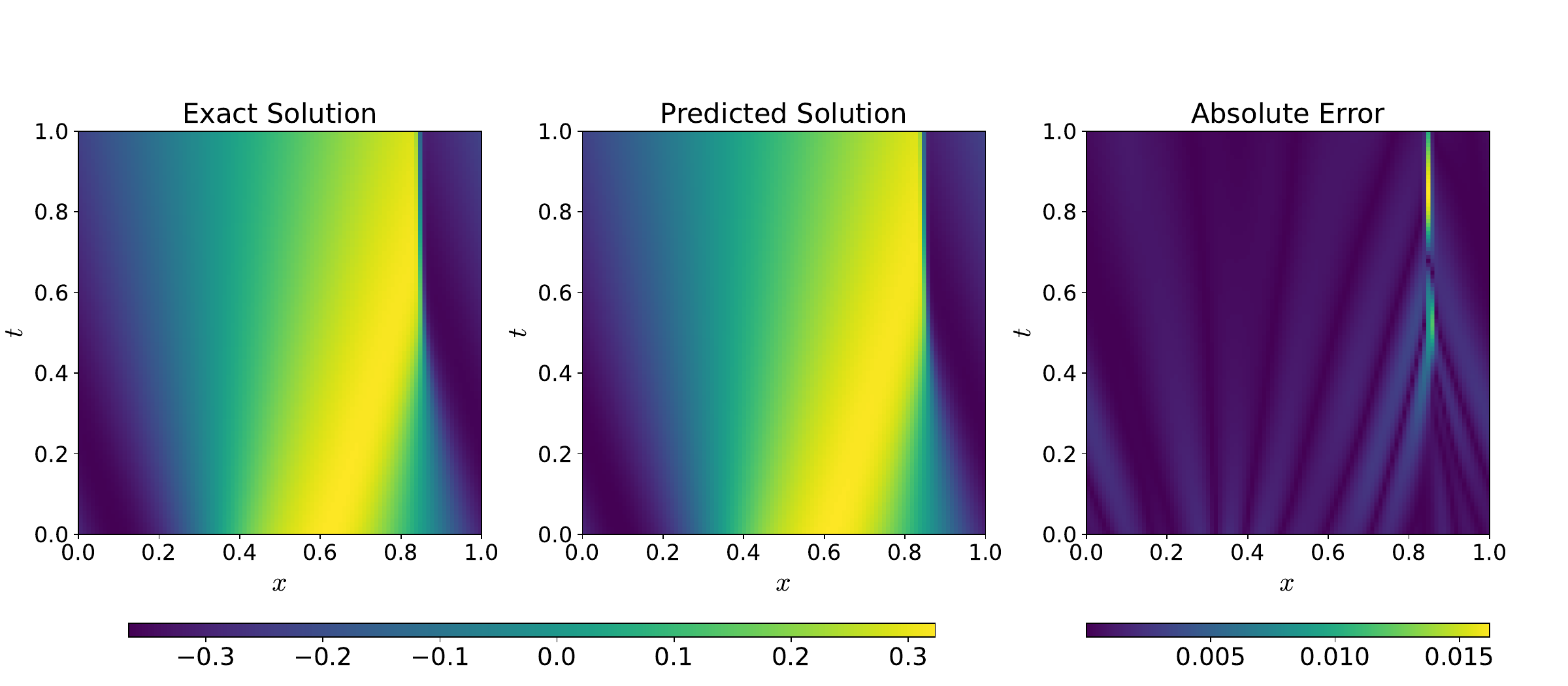}
  \caption{Burgers' equation with viscosity $10^{-3}$ (the modified DeepONet). Left to right: reference solution, model prediction, and absolute error on a representative training instance. The $L^2$ relative error is $0.61\%$.}
  \label{fig:burgers3_deeponet}
\end{figure}

\begin{figure}[H]
  \centering
  \includegraphics[width=\textwidth]{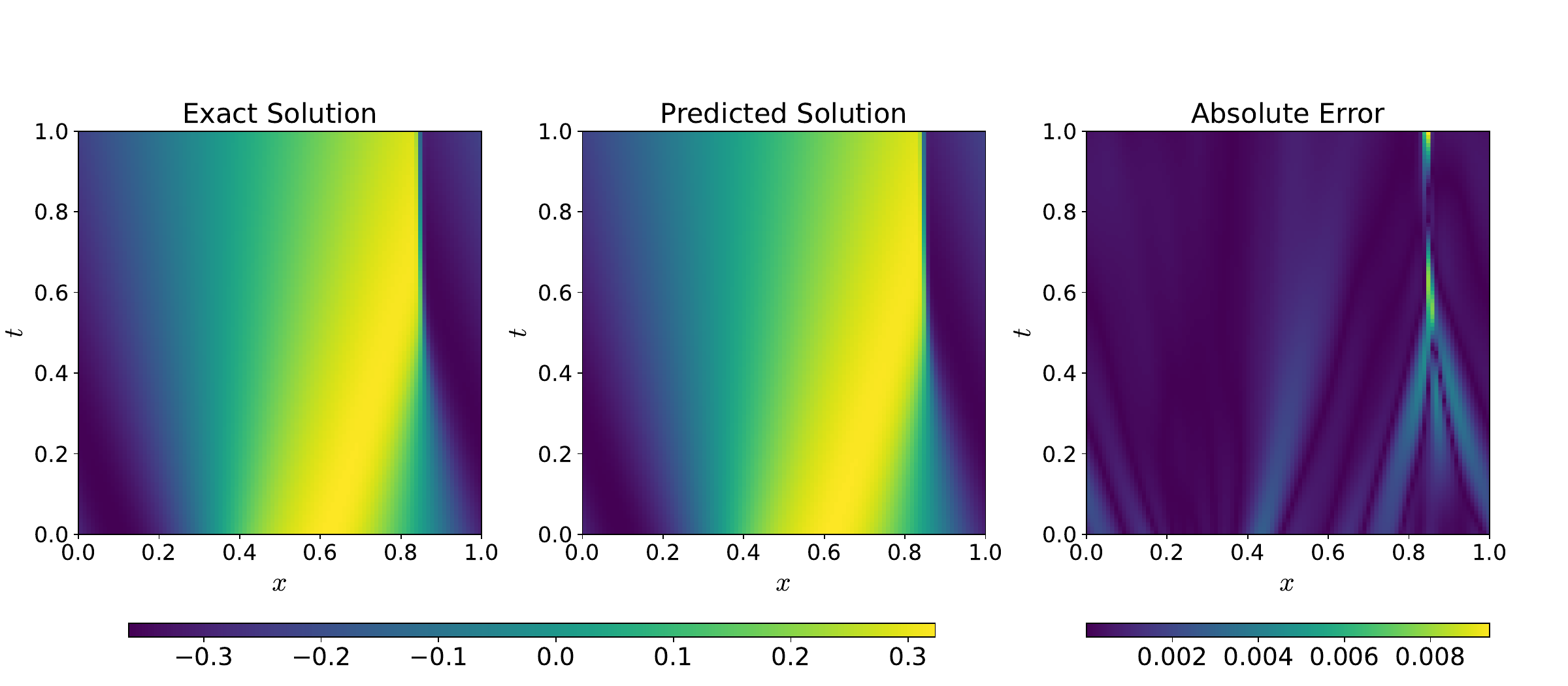}
  \caption{Burgers' equation with viscosity $10^{-3}$ (Variant TF). Left to right: reference solution, model prediction, and absolute error on the same instance as in \Cref{fig:burgers3_deeponet}. The $L^2$ relative error is $0.43\%$.}
  \label{fig:burgers3_tf}
\end{figure}

\clearpage
For Burgers' equation with viscosity $10^{-3}$, we again compare the modified DeepONet with Variant TF; see \Cref{fig:burgers3_deeponet,fig:burgers3_tf}. Variant TF achieves a slightly lower error ($0.43\%$) than the modified DeepONet ($0.61\%$) on the shown instance.

\begin{figure}[H]
  \centering
  \includegraphics[width=\textwidth]{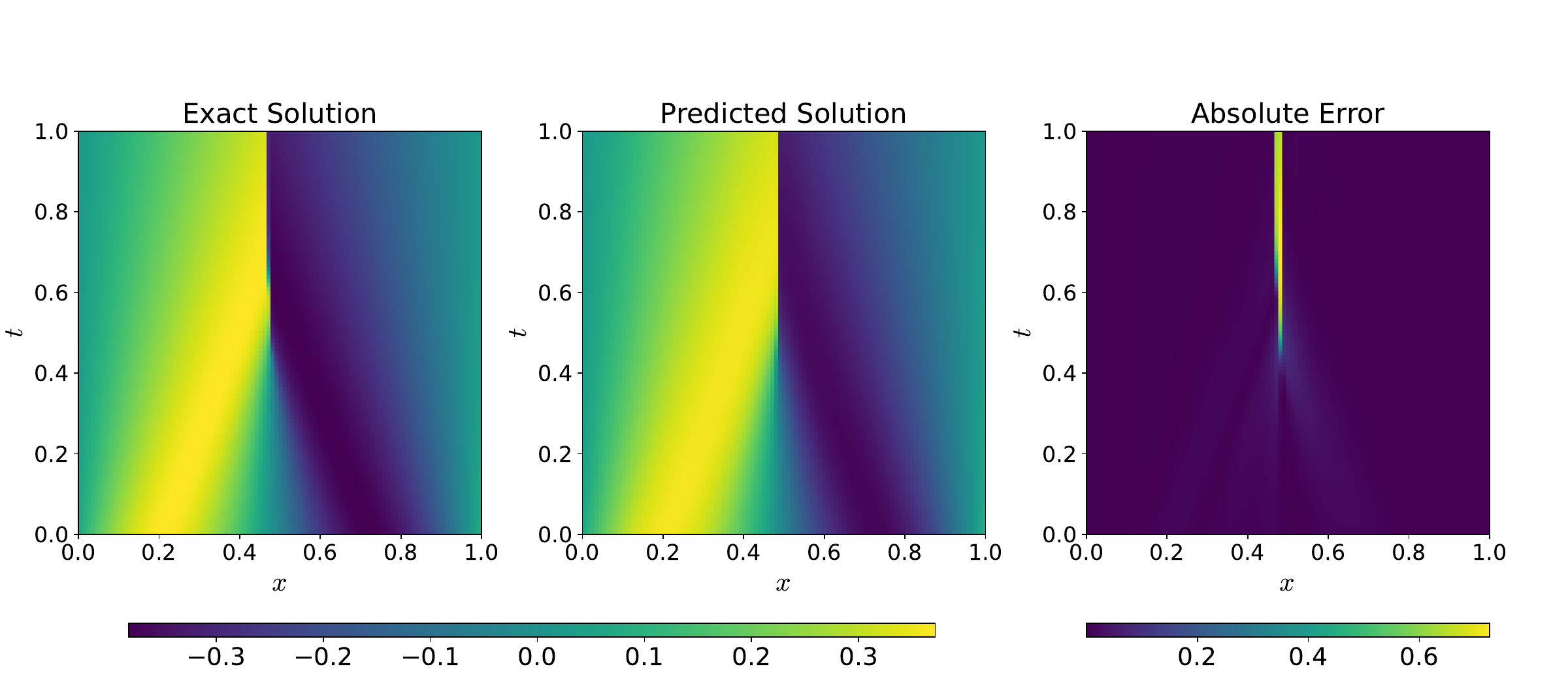}
  \caption{Burgers' equation with viscosity $10^{-4}$ (modified DeepONet). Left to right: reference solution, model prediction, and absolute error on a representative training instance. The $L^2$ relative error is $25.21\%$.}
  \label{fig:burgers4_deeponet}
\end{figure}

\begin{figure}[H]
  \centering
  \includegraphics[width=\textwidth]{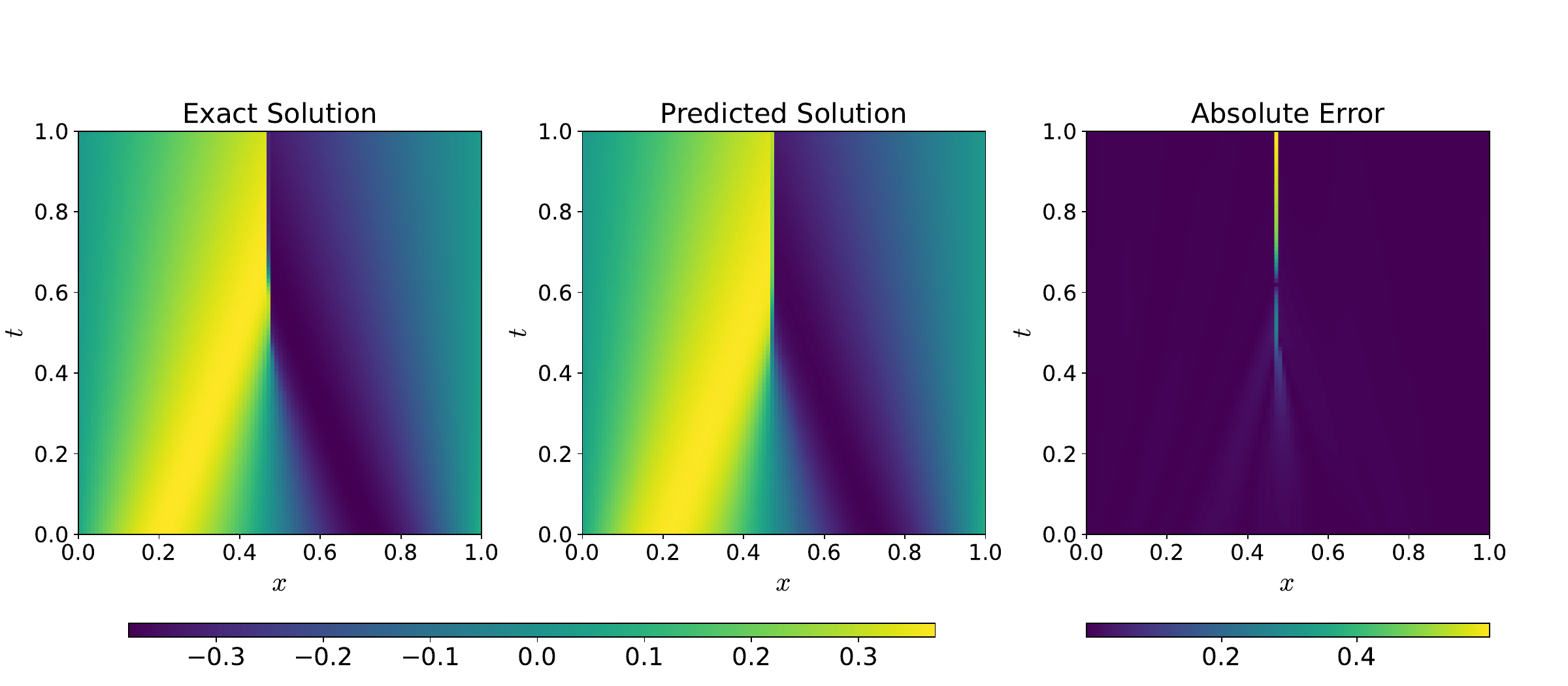}
  \caption{Burgers' equation with viscosity $10^{-4}$ (Variant TF). Left to right: reference solution, model prediction, and absolute error on the same instance as in \Cref{fig:burgers4_deeponet}. The $L^2$ relative error is $13.10\%$.}
  \label{fig:burgers4_tf}
\end{figure}

\begin{figure}[H]
  \centering
  \includegraphics[width=0.96\textwidth]{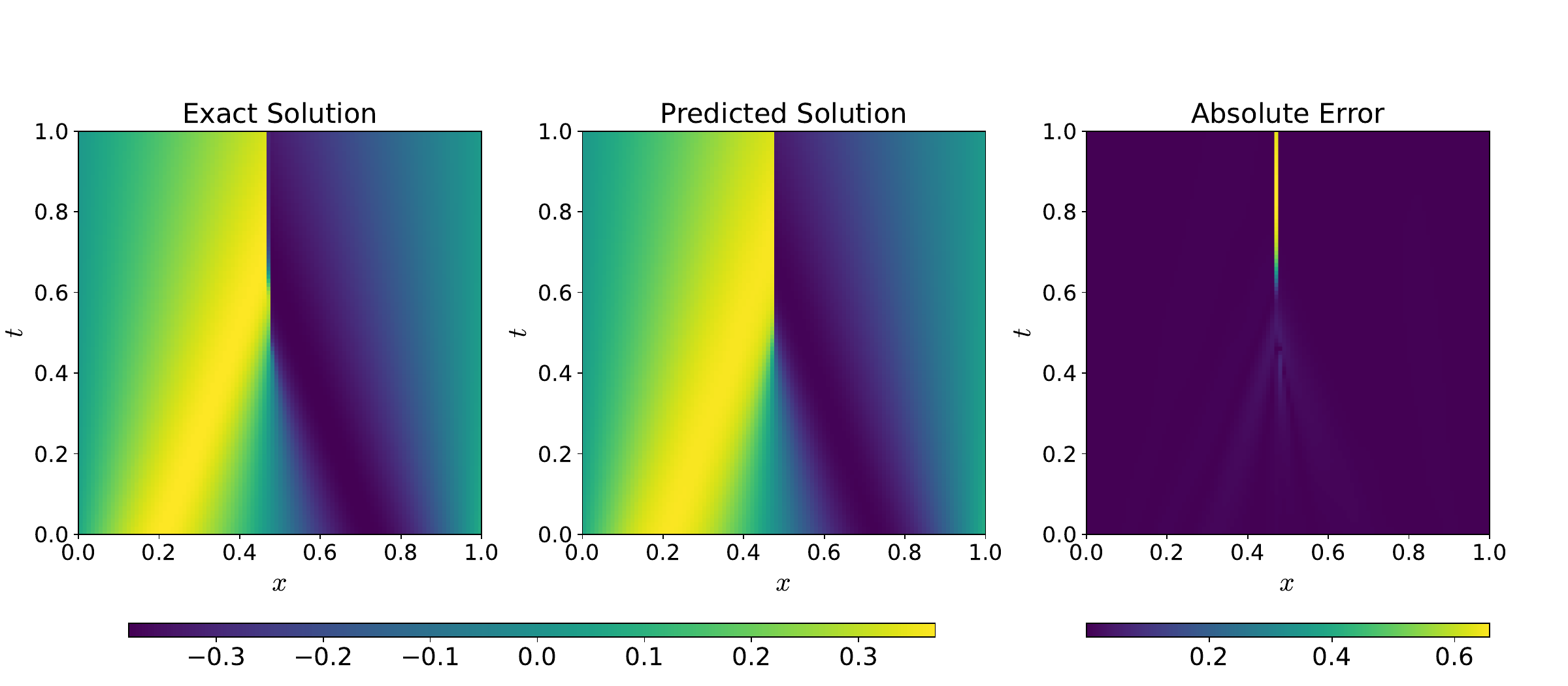}
  \caption{Burgers' equation with viscosity $10^{-4}$ (Variant BxTF). Left to right: reference solution, model prediction, and absolute error on the same instance as in \Cref{fig:burgers4_deeponet}. The $L^2$ relative error is $15.16\%$.}
  \label{fig:burgers4_bxtf}
\end{figure}

\clearpage
For Burgers' equation with viscosity $10^{-4}$, we compare the modified DeepONet with Variants TF and BxTF; see \Cref{fig:burgers4_deeponet,fig:burgers4_tf,fig:burgers4_bxtf}. Both Transformer-inspired variants outperform the modified DeepONet on this instance (relative $L^2$ errors $25.21\%$ for the modified DeepONet, $13.10\%$ for TF, and $15.16\%$ for BxTF).

\begin{figure}[H]
  \centering
  \includegraphics[width=0.96\textwidth]{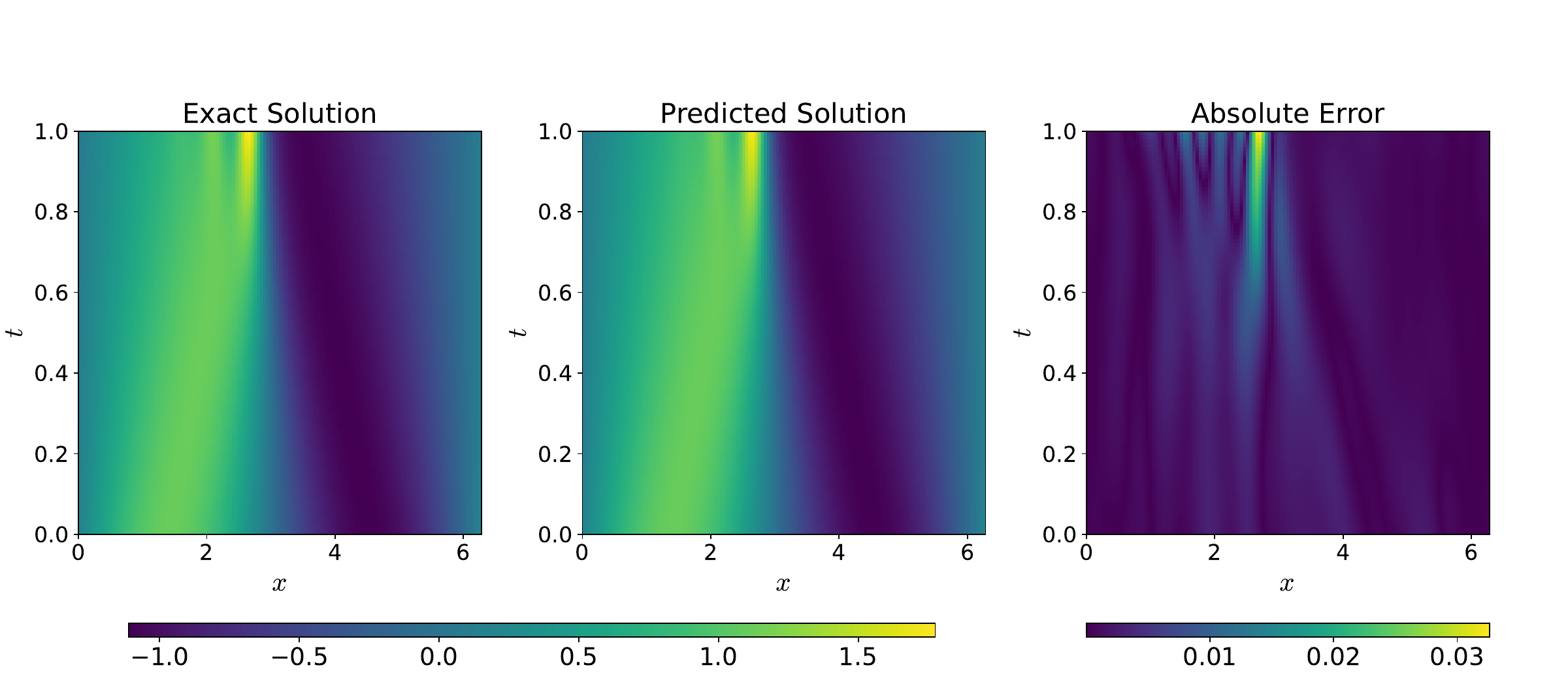}
  \caption{KdV equation (the modified DeepONet). Left to right: reference solution, model prediction, and absolute error on a representative training instance. The $L^2$ relative error is $0.44\%$.}
  \label{fig:kdv_deeponet_small}
\end{figure}

\begin{figure}[H]
  \centering
  \includegraphics[width=\textwidth]{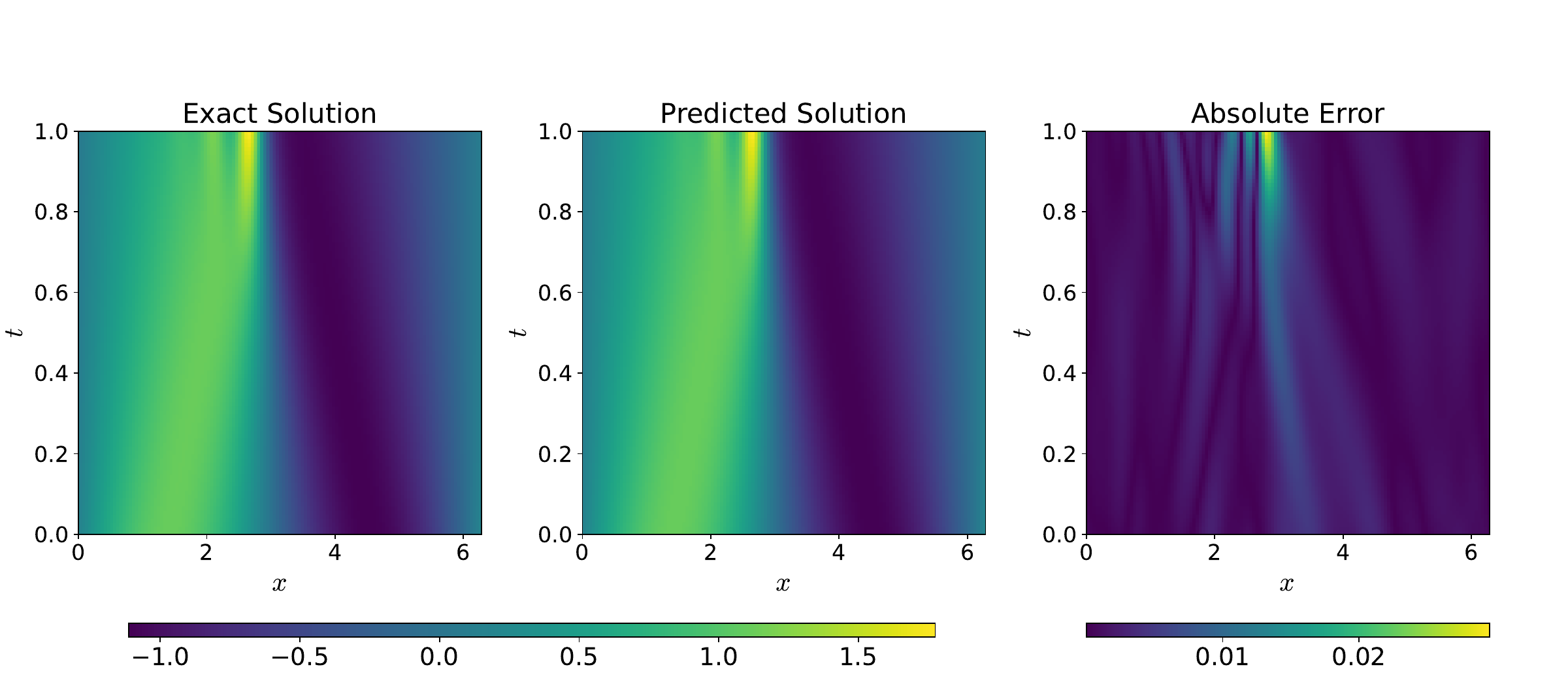}
  \caption{KdV equation (Variant BxTG). Left to right: reference solution, model prediction, and absolute error on a representative training instance. The $L^2$ relative error is $0.41\%$.}
  \label{fig:kdv_bxtg_small}
\end{figure}

\begin{figure}[H]
  \centering
  \includegraphics[width=\textwidth]{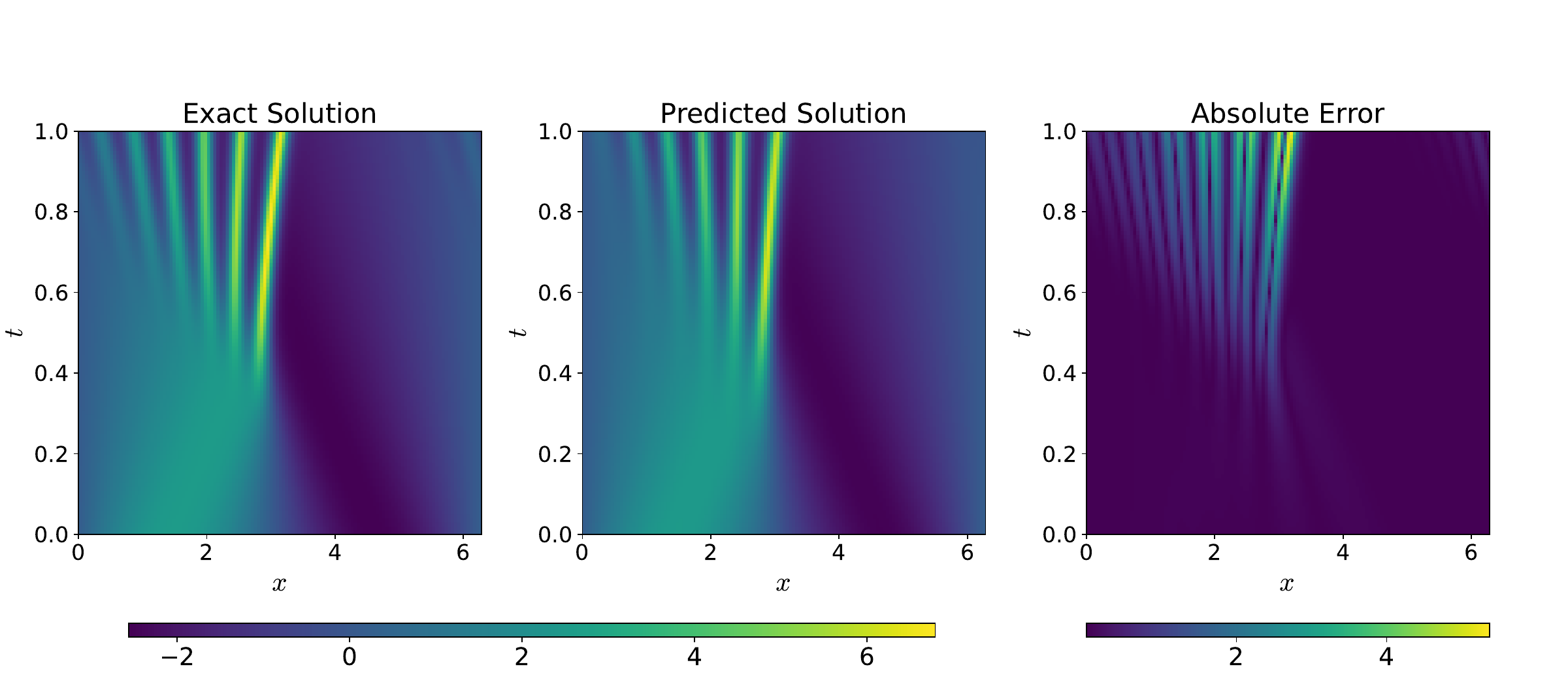}
  \caption{KdV equation (the modified DeepONet). Left to right: reference solution, model prediction, and absolute error on a representative training instance. The $L^2$ relative error is $37.96\%$.}
  \label{fig:kdv_deeponet_big}
\end{figure}

\begin{figure}[H]
  \centering
  \includegraphics[width=\textwidth]{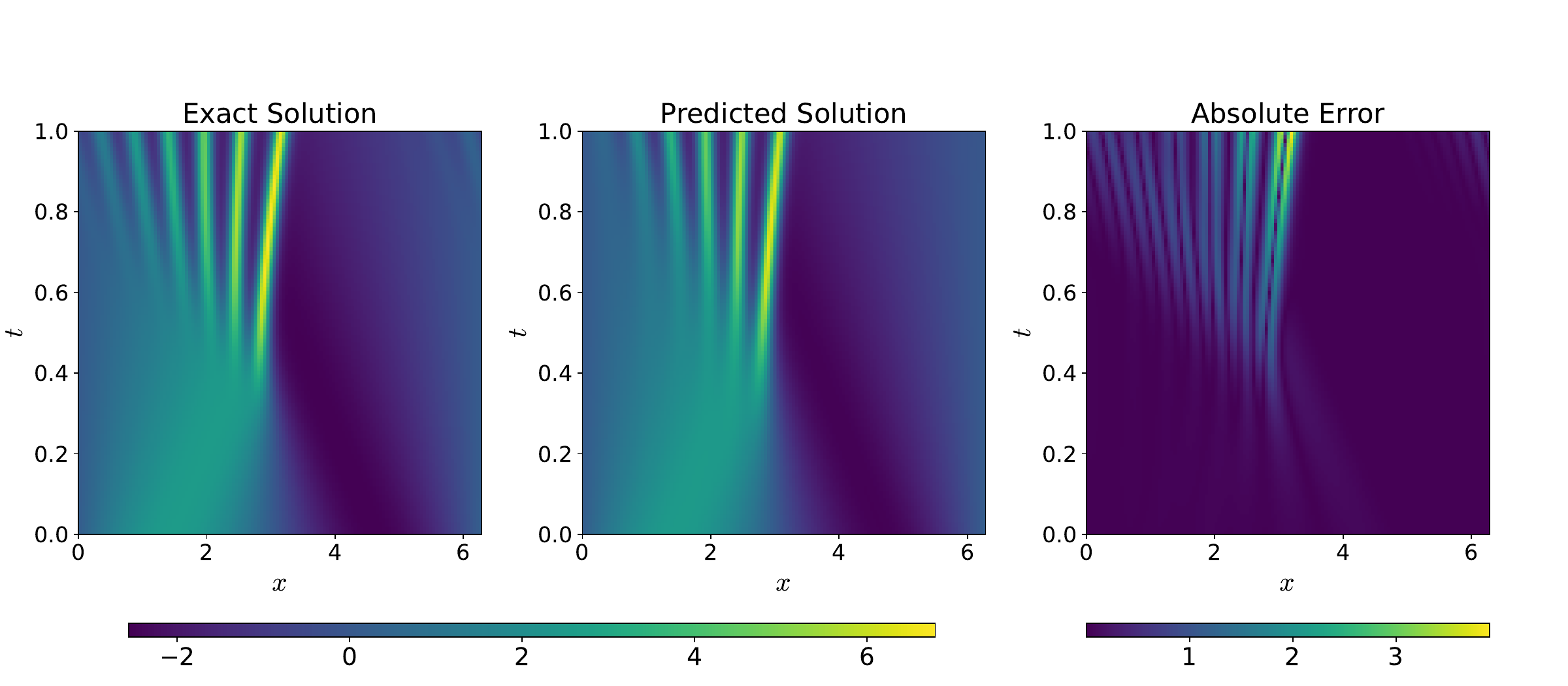}
  \caption{KdV equation (Variant BxTG). Left to right: reference solution, model prediction, and absolute error on a representative training instance. The $L^2$ relative error is $24.31\%$.}
  \label{fig:kdv_bxtg_big}
\end{figure}

For the KdV equation, we compare the modified DeepONet with the BxTG variant. \Cref{fig:kdv_deeponet_small,fig:kdv_bxtg_small} shows a representative test case of moderate difficulty on which both models achieve comparable accuracy.
In addition, as suggested by the violin plot in \Cref{fig_kdv_violin}, a small number of test cases exhibit relatively large errors. Closer inspection reveals that the errors are highly localized in regions of space-time where the exact solution exhibits large gradients; see \Cref{fig:kdv_deeponet_big,fig:kdv_bxtg_big}. On this challenging instance, Variant BxTG attains lower error (i.e., better accuracy) than the modified DeepONet.

\subsection{Results Reported in the Literature}\label{apdx23_literature_results}
In this section, we list selected results from related literature for reference and contextual comparison.

The advection equation was studied in \autocite{Wang-2022} using a physics-informed modified DeepONet. A mean $L^2$ relative error of $0.95\%$ with a standard deviation of $0.23\%$ was reported, achieved through moderate local neural tangent kernel (NTK) methods for loss term balancing. The input function distribution used in their work is the same as in our study. 

In \autocite{Wang-2021}, the authors studied a diffusion--reaction equation using a physics-informed ``vanilla'' DeepONet. Fixed weights were employed for loss term balancing, and a mean $L^2$ relative error of $0.45\%$ with a standard deviation of $0.16\%$ was reported. The input function distribution remains the same as in our study. Improved performance in our results is attributed to the use of random Fourier feature embedding.

In \autocite{Wang-2022}, Burgers' equation was studied under viscosities of $10^{-2}$, $10^{-3}$, and $10^{-4}$. The reported mean $L^2$ relative errors (with standard deviations) were $1.03\% \pm 1.35\%$, $3.69\% \pm 3.63\%$, and $7.95\% \pm 5.29\%$, respectively. For viscosities $10^{-2}$ and $10^{-3}$, our method achieved better accuracy, which we attribute to the use of deterministic Fourier feature embedding.

In \autocite{Williams-2024}, transfer learning techniques were applied to the study of both Burgers' and KdV equations using a physics-informed modified DeepONet. For the viscous Burgers' equation with viscosity $10^{-4}$, the network was initialized with pretrained parameters from the case with viscosity $10^{-3}$. This approach yielded a mean $L^2$ relative error of $7.03\%$ with a standard deviation of $4.94\%$, with moderate local NTK adaptive weights used for loss term balancing. The initial condition distribution used for the Burgers' equation matches that of our study.

For the KdV equation, the model was initialized with parameters transferred from the Burgers' equation with viscosity $10^{-4}$, which itself had been transfer-initialized from viscosity $10^{-3}$. A mean $L^2$ relative error of $3.29\%$ with standard deviation $5.94\%$ was reported, using local CK adaptive weights for balancing the loss terms. The higher error can be attributed to the increased difficulty of their initial conditions than ours. 

\section{Rigorous Statistical Analysis Procedures}\label{apdx3_data_analysis}
In this appendix, we detail the rigorous statistical analysis procedures used for a comparative evaluation of our top-performing model variants and the baseline modified DeepONet model. Recognizing that our readership may have diverse backgrounds, we have included a concise summary of the fundamental statistical theories and principles that inform our analysis, aiming to make the content self-contained and accessible.
    
For a given equation, we denote the sequences of $L^2$ relative errors for the top-performing model variant and the baseline modified DeepONet model as 
\begin{displaymath}
    \{\varepsilon_i^{\text{(v)}}\}_{i=1}^N
\end{displaymath}
and
\begin{displaymath}
    \{\varepsilon_i^{\text{(b)}}\}_{i=1}^N,
\end{displaymath}
respectively. 
Their element-wise difference sequence is 
\begin{displaymath}
    \{d_i\}_{i=1}^N
    \coloneqq
    \{\varepsilon_i^{\text{(v)}} - \varepsilon_i^{\text{(b)}}\}_{i=1}^N.
\end{displaymath}
    
\subsection{Percentage of Negative Difference}
    We first examine the element-wise difference sequence.
    Each negative term in the sequence $\{d_i\}_{i=1}^N$ indicates a test case where the variant model outperformed the baseline. Thus, the percentage of negative terms serves as a direct metric for the frequency of the variant's better accuracy across the test dataset.

\subsection{Wilcoxon Two One-Sided Tests (TOST) for Equivalence}
    To formally test for statistical equivalence in accuracy between the variant model and the baseline model, we employ the Wilcoxon two one-sided tests (TOST) procedure \autocite{Wilcoxon-1945}. This non-parametric approach assesses whether the error distributions of the variant and baseline models are effectively the same. The test is performed on the element-wise difference sequence $\{d_i\}_{i=1}^N$.

    \subsubsection{Lower-Bound Test}
    The lower-bound test determines whether the theoretical median, $M_d$, of the underlying theoretical distribution of differences $\{d_i\}_{i=1}^N$ is greater than a predefined equivalence margin, $-\Delta$. The null ($H_0$) and alternative ($H_1$) hypotheses for this test are formulated as:
    \begin{align*}
        H_0\colon M_d &\leqslant -\Delta,\\
        H_1\colon M_d &> -\Delta.
    \end{align*}
    
    The test statistic is based on the ranks of the absolute values of the differences, $\{d_i\}_{i=1}^N$. First, we establish an unsigned rank sequence, $\{\widehat{R}_i\}_{i=1}^N$, where the rank $\widehat{R}_i=1$ is assigned to the difference with the smallest absolute value, and $\widehat{R}_i=N$ is assigned to the one with the largest absolute value.
    
    Next, we construct the signed-rank sequence $\{R_i\}_{i=1}^N$. A sign is assigned to each unsigned rank $\widehat{R}_i$ based on its corresponding difference $d_i$: if $d_i > -\Delta$, we set $R_i = +\widehat{R}_i$; otherwise, we set $R_i = -\widehat{R}_i$. From this sequence, we define the positive and negative rank-sum statistics as:
    \begin{displaymath}
        w^+ \coloneqq \sum_{R_i>0} R_i
        \quad \text{and}\quad
        w^- \coloneqq \sum_{R_i\leqslant 0} R_i.
    \end{displaymath}

    The objective is to calculate the $p$-value, $p^{(\text{lower})}$, defined as the probability of observing a positive (stochastic) rank-sum statistic $W^+$ at least as large as the measured $w^+$, under the assumption that the null hypothesis $H_0$ is true. Calculating this probability directly under the composite hypothesis $H_0$ (i.e., $M_d \leqslant -\Delta$) is complex. We therefore simplify the problem by first analyzing the boundary case of the null hypothesis, $\widetilde{H}_0 \colon M_d = -\Delta$. Under this simpler condition, we calculate a corresponding probability, $\widetilde{p}^{(\text{lower})}$. This approach, however, hinges on understanding the stochastic nature of $W^+$, whose source of randomness has not yet been defined. We will first clarify this, and then demonstrate that $\widetilde{p}^{(\text{lower})}$ provides a valid $p$-value for the original hypothesis test.
    
    The source of randomness for the test statistic $W^+$ arises from the assumption of symmetry inherent in the null hypothesis. Specifically, under the boundary condition $\widetilde{H}_0$ (where the median $M_d = -\Delta$), any difference $d_i$ is equally likely to fall above or below $-\Delta$. Consequently, when constructing the signed-rank sequence $\{R_i\}_{i=1}^N$, the assignment of a positive or negative sign to each corresponding unsigned rank $\widehat{R}_i$ is an equiprobable event. This generates a sample space of $2^N$ possible, equally likely signed-rank sequences. The random variable $W^+$ is thus defined as the positive rank-sum calculated over this discrete uniform probability space.
    
    We now connect the probability calculated in the boundary case, $\widetilde{p}^{(\text{lower})}$, to the $p$-value for the original composite hypothesis, $p^{(\text{lower})}$. The boundary case $\widetilde{H}_0$ represents the ``edge'' of the parameter space defined by $H_0$. For any other case within $H_0$ (i.e., where $M_d < -\Delta$), the distribution of $\{d_i\}_{i=1}^N$ is shifted further to the left. This implies that the probability of observing a difference greater than $-\Delta$ is lower than it is under the boundary condition. Consequently, the probability of achieving a given positive rank-sum $w^+$ is maximized under $\widetilde{H}_0$. This gives us the following inequality:
    \begin{displaymath}
        \sup_{M_d \leqslant -\Delta} \Pleft{W^+ \geqslant w^+} = \Psubleft{\widetilde{H}_0}{W^+ \geqslant w^+}.
    \end{displaymath}
    
    By definition, the $p$-value is the maximum probability of observing a result at least as extreme as the one measured, over the entire parameter space of the null hypothesis. Based on the reasoning above, this maximum occurs at the boundary. Therefore, we can calculate the $p$-value for the composite hypothesis $H_0$ by using the simpler boundary case $\widetilde{H}_0$:
    \begin{displaymath}
        p^{(\text{lower})}
        \coloneqq
        \sup_{M_d \leqslant -\Delta} \Pleft{W^+ \geqslant w^+}
        =
        \Psubleft{\widetilde{H}_0}{W^+ \geqslant w^+}
        =
        \widetilde{p}^{(\text{lower})}.
    \end{displaymath}
    A small $p^{(\text{lower})}$ indicates that the observed data is unlikely if the null hypothesis were true. Following standard statistical convention, if $p^{(\text{lower})}$ is below a pre-specified significance level (e.g., $0.05$), we reject $H_0$ and adopt the alternative hypothesis, $H_1$.
    
    \subsubsection{Upper-Bound Test}
    Symmetrically, the upper-bound test determines whether the theoretical median of the underlying theoretical distribution of differences, $M_d$, is less than a predefined equivalence margin, $+\Delta$. The null ($H_0$) and alternative ($H_1$) hypotheses are formulated as:
    \begin{align*}
        H_0\colon M_d &\geqslant +\Delta,\\
        H_1\colon M_d &< +\Delta.
    \end{align*}
    The corresponding $p$-value, $p^{(\text{upper})}$, is derived using a procedure analogous to that of the lower-bound test, but by testing against the upper margin.

    \subsubsection{Conclusion for Equivalence}
    The two one-sided tests are combined to reach a conclusion on equivalence. If both the lower-bound and upper-bound tests are statistically significant (i.e., $p^{(\text{lower})} < 0.05$ and $p^{(\text{upper})} < 0.05$), we reject both corresponding null hypotheses. This allows us to conclude that the theoretical median of the difference distribution, $M_d$, lies within the equivalence interval $(-\Delta, +\Delta)$. In practical terms, this outcome provides statistical evidence that the variant and baseline models are equivalent in prediction accuracy as reflected by their $L^2$ relative errors.
    
    In practice, determining an appropriate equivalence margin $\Delta$ is context-dependent and often requires a balance between statistical rigor and practical interpretability (cf.~\cite{Schuirmann-1987,Wellek-2010}). In analogy to the use of relative thresholds in equivalence testing (e.g., the $20\%$ rule in bioequivalence studies), we set the equivalence margin as one fifth of the minimal error over all test cases given by the baseline model, i.e., 
    \begin{displaymath}
        \Delta \coloneq 0.2 \times \min_i\varepsilon_i^{\text{(b)}}.
    \end{displaymath}
    This choice ensures that the equivalence margin is (1) anchored to the performance scale of the baseline, (2) interpretable as a practically negligible difference, and (3) objectively determined from the data rather than set arbitrarily and subjectively.

\subsection{Determining the Model with Better Accuracy if Inequivalent}
If the TOST procedure does not establish equivalence, we proceed to determine which model exhibits better accuracy. In this scenario, the theoretical median of the underlying theoretical distribution of the differences, $M_d$, is presumed to lie outside the equivalence interval $(-\Delta, +\Delta)$. We use the sample median of the observed differences, denoted $\widehat{M}_d$, as an estimator for the theoretical median. The sign of the sample median indicates better model:
\begin{itemize}
    \item If $\widehat{M}_d > 0$, the differences tend to be positive, suggesting that the baseline model has lower prediction errors and thus better accuracy.
    \item If $\widehat{M}_d < 0$, the differences tend to be negative, suggesting that the variant model has lower prediction errors and therefore outperforms the baseline in prediction accuracy.
\end{itemize}

\subsection{Quantifying the Accuracy Difference}
If the models are found to be inequivalent, we next quantify the magnitude of the difference of errors. For this purpose, we employ a non-parametric effect size measure analogous to Glass's $\Delta$ \autocite{Glass-1976,Wilcox-2011}. This metric standardizes the difference of errors by scaling it against the variability of the baseline model.

The numerator is the sample median of the error difference sequence
\begin{displaymath}
    \widehat{M}_d = \text{median}(\{d_i\}_{i=1}^N).
\end{displaymath}
This value represents the typical accuracy gap between the variant model and the baseline model, with its sign indicating which model has better accuracy.

The denominator is the Median Absolute Deviation (MAD) of the baseline model's error sequence, $\{\varepsilon_i^{\text{(b)}}\}_{i=1}^N$. The MAD is a robust measure of statistical dispersion. Denoting the median of the baseline errors as 
$\widehat{M}_\varepsilon = \text{median}(\{\varepsilon_k^{\text{(b)}}\}_{k=1}^N)$, 
the MAD is defined as:
\begin{displaymath}
    \text{MAD}
    \coloneqq
    \text{median}
    \Bigl(
    \bigl\{
    \abs{\varepsilon_i^{\text{(b)}} - \widehat{M}_\varepsilon}
    \bigr\}_{i=1}^N 
    \Bigr).
\end{displaymath}
Thus, our effect size measure, the ratio of the median difference to the baseline's MAD,
\begin{displaymath}
    \text{Effect Size} = \frac{\widehat{M}_d}{\text{MAD}},
\end{displaymath}
quantifies the magnitude of the accuracy difference relative to the baseline's own variability. This provides a standardized and interpretable metric of the accuracy difference.

\subsection{Behavioral Correlation of the Models}
Finally, to assess the behavioral similarity between the variant model and the baseline modified DeepONet model, we compute Spearman's rank correlation coefficient $\rho$ \autocite{Spearman-1904,Hollander-2013}. This non-parametric measure evaluates the strength and direction of the monotonic relationship between the two models' error profiles.

Given the error sequences $\{\varepsilon_i^{\text{(v)}}\}_{i=1}^N$ and $\{\varepsilon_i^{\text{(b)}}\}_{i=1}^N$, each is converted into its corresponding rank sequence, denoted $\{\widehat{R}_i^{\text{(v)}}\}_{i=1}^N$ and $\{\widehat{R}_i^{\text{(b)}}\}_{i=1}^N$. Spearman's $\rho$ is then calculated as the Pearson correlation coefficient applied to these rank sequences. A high positive correlation ($\rho \approx 1$) indicates that the models tend to perform well or poorly on the same test cases, suggesting similar underlying behavior. 

\begin{landscape}
\subsection{Statistical Analysis Results}
The results of the rigorous statistical analysis process outlined earlier in this section are summarized in \Cref{tab_variant_statistics}.
\begin{table}[H]
\centering
\caption{Statistical comparison of selected variants against the modified DeepONet across different equations.}
\label{tab_variant_statistics}
\setlength{\tabcolsep}{8pt} 
\renewcommand{\arraystretch}{1.2} 
\begin{tabular}{lcccc}
\toprule
\textbf{Equation}            & \textbf{Advection} & \textbf{Diffusion--Reaction} & \textbf{Burgers' ($\nu=10^{-2}$)} & \textbf{Burgers' ($\nu=10^{-3}$)} \\
\midrule
\textbf{Variant}             & BxTG               & TL                           & TF                               & TF                               \\
\midrule
Percentage (\%)              & $57.00$            & $64.90$                      & $20.40$                          & $47.80$                          \\
Equivalence Margin (\%)      & $0.10$             & $0.02$                       & $0.01$                           & $0.06$                           \\
$p^{(\text{lower})}$         & $0.00$             & $0.0027$                     & $0.00$                           & $0.00$                           \\
$p^{(\text{upper})}$         & $0.00$             & $0.00$                       & $1.00$                           & $0.00$                           \\
Equivalent or Not            & Yes                & Yes                          & No                               & Yes                              \\
Median Difference (\%)       & $-0.0063$          & $-0.0127$                    & $0.0248$                         & $0.0115$                         \\
Glass's $\Delta$             & --                 & --                           & $0.397$                          & --                               \\
Spearman's $\rho$            & $0.993$            & $0.769$                      & $0.858$                          & $0.869$                          \\
\toprule
\textbf{Equation}            & \multicolumn{2}{c}{\textbf{Burgers' ($\nu=10^{-4}$)}} & \multicolumn{1}{c}{\textbf{KdV}} \\
\midrule
\textbf{Variant}             & TF                 & BxTF                         & BxTG                             \\
\midrule
Percentage (\%)              & $68.80$            & $83.00$                      & $56.00$                          \\
Equivalence Margin (\%)      & $0.86$             & $0.86$                       & $0.04$                           \\
$p^{(\text{lower})}$         & $1.00$             & $1.00$                       & $0.823$                          \\
$p^{(\text{upper})}$         & $0.00$             & $0.00$                       & $0.00$                           \\
Equivalent or Not            & No                 & No                           & No                               \\
Median Difference (\%)       & $-3.3255$          & $-4.7989$                    & $-0.0152$                        \\
Glass's $\Delta$             & $-0.393$           & $-0.568$                     & $-0.054$                         \\
Spearman's $\rho$            & $0.515$            & $0.542$                      & $0.860$                          \\
\bottomrule
\end{tabular}
\end{table}
\thispagestyle{empty}
\end{landscape} 

\clearpage
\section{Symbols \& Notation}
    \Cref{tab:notation} summarizes the main symbols and notation used in this work.
    \begin{table}[h]
    \centering
    \caption{Summary of the main symbols and notation used in this work.}
    \label{tab:notation}
    \begin{tabular}{l p{0.58\linewidth}}
    \toprule
    \textbf{Notation} & \textbf{Description} \\
    \midrule
    $t$                           & Temporal query coordinate\\
    $n_t$                         & Number of time steps in solution dataset\\
    $\bm{x}\in\mathbb{R}^d$       & Spatial query coordinate\\
    $n_x$                         & Number of spatial points in solution dataset\\
    $\bm{y}$                      & Trunk input of ``vanilla'' DeepONet\\
    $\bm{u}$                      & Input function (branch input of ``vanilla'' DeepONet)\\
    $\{\bm{x}_i\}_{i=1}^m$        & $m$ sensor spatial locations of input functions $\bm{u}$\\
    $\widehat{\bm{u}}_{\Lambda} = (\hat{u}_{\bm{k}})_{\bm{k} \in \Lambda}$ & A truncated set of Fourier coefficients of $\bm{u}$\\
    $\bm{s}$                      & A solution to a parametric PDE\\
    $G$                           & Solution operator\\
    $G^{\theta}$                  & Approximation of a solution operator\\
    $\theta$                      & All trainable parameters\\
    PDE                           & Partial differential equation\\
    GRF                           & Gaussian random field\\
    NTK                           & Neural tangent kernel\\
    CK                            & Conjugate kernel\\
    BRDR                          & balanced-residual-decay-rate\\
    $(b_1, b_2, \ldots, b_w)^\intercal \in \RNS^w$ & Branch output\\
    $(\gamma_1, \gamma_2, \ldots, \gamma_w)^\intercal \in \RNS^w$ & Trunk output\\
    $D$                           & Diffusion coefficient of Diffusion--Reaction equation\\
    $k$                           & Reaction rate of Diffusion--Reaction equation\\
    $\nu$                         & Viscosity of Burgers' equation\\
    $\delta$                      & Dispersion parameter of KdV equation\\
    \bottomrule
    \end{tabular}
    \end{table}

\end{appendices}

\section*{References}
\addcontentsline{toc}{section}{\protect\numberline {}{References}}
\printbibliography[heading=none]  
\end{document}